\documentclass[10pt,twocolumn,letterpaper]{article}

\usepackage{cvpr}              %

\usepackage[dvipsnames]{xcolor}

\definecolor{cvprblue}{rgb}{0.21,0.49,0.74}
\usepackage[pagebackref,breaklinks,colorlinks,citecolor=cvprblue]{hyperref}
\usepackage{duckuments}
\usepackage{color,xcolor}
\usepackage{epsfig}
\usepackage{graphicx}

\usepackage{microtype}
\usepackage[font=small]{caption}
\usepackage{arydshln}
\usepackage{tabularx}
\usepackage{adjustbox}
\usepackage{array}
\usepackage{booktabs}
\usepackage{colortbl}
\usepackage{float,wrapfig}
\usepackage{hhline}
\usepackage{multirow}
\usepackage{subcaption} %
\usepackage[percent]{overpic}

\usepackage{breqn}

\usepackage{duckuments}
\usepackage{amsmath}

\usepackage{bm}
\usepackage{nicefrac}
\usepackage{microtype}
\usepackage{dsfont}
\usepackage{changepage}
\usepackage{extramarks}
\usepackage{fancyhdr}
\usepackage{setspace}
\usepackage{soul}
\usepackage{xspace}
\usepackage{hhline}
\usepackage{algorithmicx}
\usepackage{algpseudocode}
\usepackage{pifont}
\usepackage{booktabs}
\usepackage{multirow}

\usepackage{makecell}

\usepackage{enumitem}
\usepackage[title]{appendix}

\def\Y{\mathcal{Y}} %

\def\W{{\mathbf{W}}} %
\def\c{{\mathbf{c}}}
\def\x{{\mathbf{x}}}
\def\y{{\mathbf{y}}}
\def\z{{\mathbf{z}}}

\def\c{{\mathbf{c}}}

\def\z{{\mathbf{z}}}

\newcommand{\smallsim}{\smallsym{\mathrel}{\sim}\hspace{-.5mm}}

\makeatletter
\newcommand{\smallsym}[2]{#1{\mathpalette\make@small@sym{#2}}}
\newcommand{\make@small@sym}[2]{%
  \vcenter{\hbox{$\m@th\downgrade@style#1#2$}}%
}
\newcommand{\downgrade@style}[1]{%
  \ifx#1\displaystyle\scriptstyle\else
    \ifx#1\textstyle\scriptstyle\else
      \scriptscriptstyle
  \fi\fi
}
\makeatother

\newcommand*{\menlo}{\fontfamily{lmtt}\fontsize{8}{8}\selectfont }

\newcommand{\nupur}[1]{\textcolor{black}{#1}}

\newcommand\blfootnote[1]{%
  \begingroup
  \renewcommand\thefootnote{}\footnote{#1}%
  \addtocounter{footnote}{-1}%
  \endgroup
}

\newcommand{\reffig}[1]{Figure~\ref{fig:#1}}
\newcommand{\refsec}[1]{Section~\ref{sec:#1}}
\newcommand{\refapp}[1]{Appendix~\ref{sec:#1}}
\newcommand{\reftbl}[1]{Table~\ref{tbl:#1}}

\newcommand{\refeq}[1]{Eqn.~\ref{eq:#1}}

\newcommand{\lblfig}[1]{\label{fig:#1}}
\newcommand{\lblsec}[1]{\label{sec:#1}}

\newcommand{\ignorethis}[1]{}

\newcommand{\myparagraph}[1]{\vspace{1pt} \noindent \textbf{#1} \ }

\def\1{\bm{1}}

\newcolumntype{L}[1]{>{\raggedright\let\newline\\\arraybackslash\hspace{0pt}}m{#1}}
\newcolumntype{C}[1]{>{\centering\let\newline\\\arraybackslash\hspace{0pt}}m{#1}}
\newcolumntype{R}[1]{>{\raggedleft\let\newline\\\arraybackslash\hspace{0pt}}m{#1}}

\newcommand{\ignore}[1]{}

\renewcommand*{\thefootnote}{\arabic{footnote}}

\makeatletter
\DeclareRobustCommand\onedot{\futurelet\@let@token\@onedot}
\def\@onedot{\ifx\@let@token.\else.\null\fi\xspace}

\def\etal{\emph{et al}\onedot}
\makeatother

\def\paperID{*****} %
\def\confName{CVPR}
\def\confYear{2024}

\title{Customizing Text-to-Image Diffusion with Object Viewpoint Control}

\author{
Nupur Kumari$^{1}$\thanks{Equal contribution} \hspace{10mm} Grace Su$^{1}$$^{\ast}$ \hspace{10mm} Richard Zhang$^{2}$ \\
Taesung Park$^{2}$ \hspace{10mm}  Eli Shechtman$^{2}$ \hspace{10mm} Jun-Yan Zhu$^{1}$\\ \\
$^{1}$Carnegie Mellon University \hspace{10mm} 
$^{2}$Adobe Research
}

\begin{document}

\twocolumn[{%
\renewcommand\twocolumn[1][]{#1}%
\maketitle

\begin{center}
    \centering
    \includegraphics[width=\linewidth]{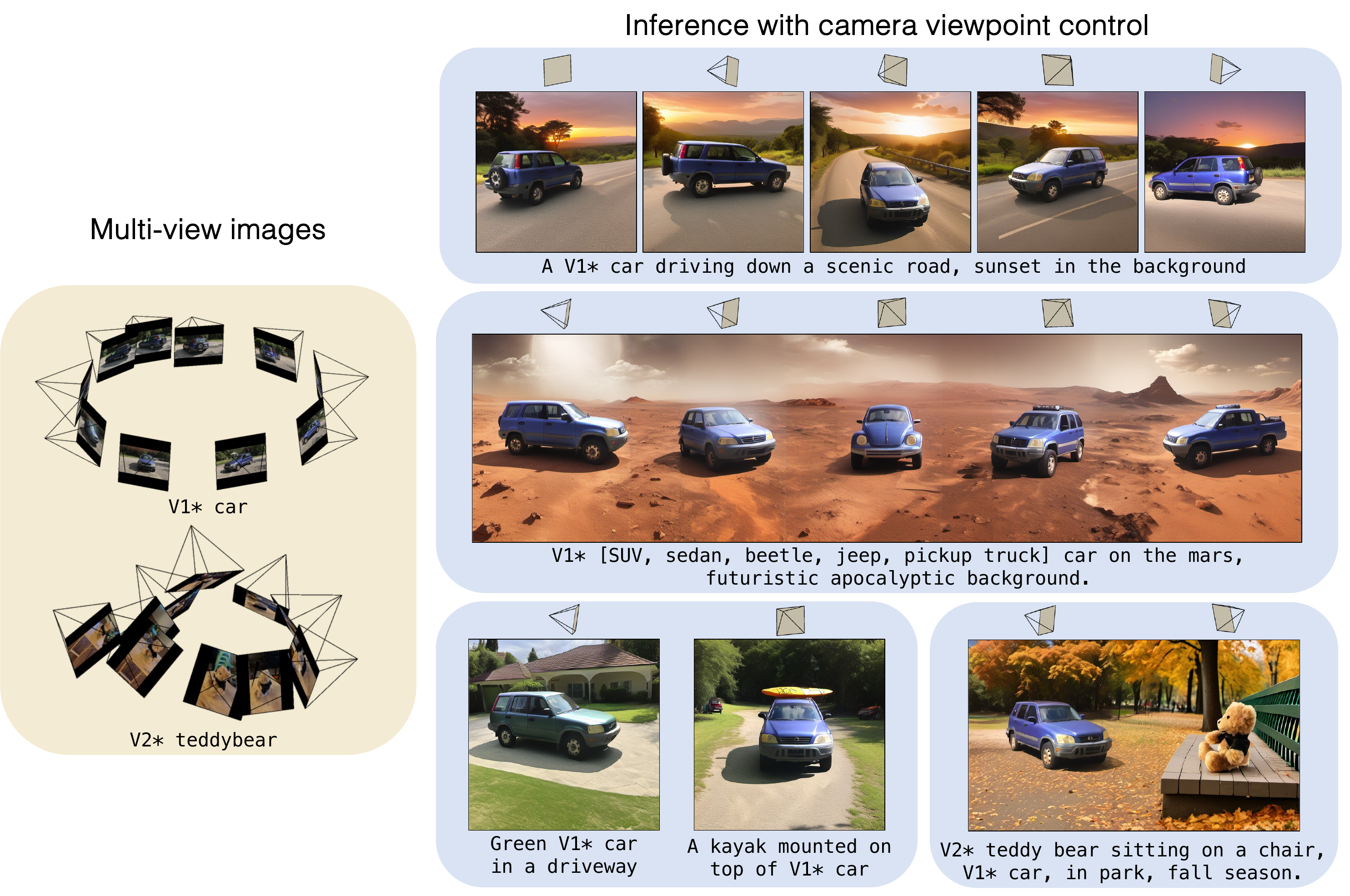}
    \captionof{figure}{Given multi-view images of a new object (left), denoted as {\menlo V$^*$ <category name>}, we create a customized text-to-image diffusion model with object viewpoint control. The customized model allows users to specify the target viewpoint for the object while synthesizing it in novel appearances and scenes, such as {\menlo A green V$^*$ car},  or {\menlo A beetle-like V$^*$ car}. We can also generate panorama images or compose multiple concepts while controlling each object's viewpoint by using MultiDiffusion~\cite{bar2023multidiffusion} with our model. 
    }
    \label{fig:teaser}
\end{center}
}]
\maketitle

\begin{abstract}
\vspace{-10pt}
 \blfootnote{\small{* indicates equal contribution}}
Model customization introduces new concepts to existing text-to-image models, enabling the generation of these new concepts/objects in novel contexts.
However, such methods lack accurate camera view control with respect to the new object, and users must resort to prompt engineering (e.g., adding ``top-view'') to achieve coarse view control. In this work, we introduce a new task -- enabling explicit control of the \emph{object viewpoint} in the customization of text-to-image diffusion models. This allows us to modify the custom object's properties and generate it in various background scenes via text prompts, all while incorporating the object viewpoint as an additional control. This new task presents significant challenges, as one must harmoniously merge a 3D representation from the multi-view images with the 2D pre-trained model. To bridge this gap, we propose to condition the diffusion process on the 3D object features rendered from the target viewpoint. During training, we fine-tune the 3D feature prediction modules to reconstruct the object's appearance and geometry, while reducing overfitting to the input multi-view images. Our method outperforms existing image editing and model customization baselines in preserving the custom object's identity while following the target object viewpoint and the text prompt.

\end{abstract}
    
\section{Introduction}
Recently, we have witnessed an explosion of works on customizing text-to-image models~\cite{ruiz2022dreambooth,gal2022image,kumari2023multi,chen2023subject}. Such methods enable a model to quickly acquire visual concepts, such as personal objects and favorite places, and re-imagine them with new environments and attributes. For instance, we can customize a model on our teddy bear and prompt it with ``Teddy bear on a bench in the park.'' However, these methods lack precise viewpoint control, as the pre-trained model is trained purely on 2D images without ground truth camera poses. As a result, users often rely on text prompts such as ``front-facing'' or ``side-facing'', a tedious and unwieldy process to control views. 

What if a user wishes to control the custom object's viewpoint while synthesizing it in a different context, e.g., the car in \reffig{teaser}? In this work, we introduce a new task: given multi-view images of an object, we customize a text-to-image model while enabling control of the object's viewpoint. During inference, our method offers the flexibility of conditioning the generation process on both a target viewpoint and a text prompt.

Neural rendering methods have allowed us to accurately control the 3D viewpoint of an \textit{existing} scene, given multi-view images~\cite{kerbl20233d,muller2022instant,barron2023zip,barron2021mip}. Similarly, we seek to imagine the object from novel views but in a \textit{new} context. However, as pre-trained diffusion models, such as Latent Diffusion models~\cite{rombach2022high}, are built upon a purely 2D representation, connecting the 3D neural representation of the object to the 2D internal features of the diffusion model remains challenging.

In this work, we introduce CustomDiffusion360, a new method to bridge the gap between 3D neural capture and 2D text-to-image diffusion models by providing viewpoint control for custom objects. More concretely, given multi-view images of an object, we introduce FeatureNeRF blocks in the diffusion model U-Net's intermediate feature spaces to learn view-dependent features. To condition the generation process on a target viewpoint, we render the FeatureNeRF output from this viewpoint and merge it with the diffusion features using linear projection layers. We only train the new linear projection layers and FeatureNeRF blocks, added to a subset of transformer layers, to preserve object identity while maintaining generalization. The pre-trained model's parameters remain frozen, thus keeping our method computationally and storage efficient.

We build our method on Stable Diffusion-XL~\cite{podell2023sdxl} and show results on various object categories, such as cars, chairs, motorcycles, teddy bears, and toys. We compare our approach with image editing~\cite{meng2021sdedit,brooks2023instructpix2pix}, model customization~\cite{hu2021lora}, and NeRF editing methods~\cite{dong2024vica}. Our method achieves high alignment with the custom object's identity and target viewpoint while adhering to the user-provided text prompt. We show that integrating the 3D object information into the text-to-image model, as done by our method, enhances performance over 2D and 3D editing baseline methods. Additionally, our method can be combined with other algorithms~\cite{meng2021sdedit,bar2023multidiffusion} for applications such as object viewpoint adjustment in the same background, panorama synthesis, and object composition. %

\section{Related Works}

\myparagraph{Text-based image synthesis.}
Large-scale text-to-image models~\cite{saharia2022photorealistic,ramesh2022hierarchical,gafni2022make,kang2023scaling,yu2022scaling} have become ubiquitous for generating photorealistic images from text prompts. This progress has been driven by the availability of large-scale datasets~\cite{schuhmann2021laion} as well as advancements in model architecture and training  objectives~\cite{dhariwal2021diffusion,sauer2023stylegan,karras2022elucidating,peebles2023scalable,karras2023analyzing}. Among them, diffusion models~\cite{song2020denoising,ho2020denoising} have emerged as a powerful family of models that generate images by gradually denoising Gaussian noise.

\myparagraph{Image editing with text-to-image diffusion.} 
One of the first works, SDEdit~\cite{meng2021sdedit}, exploited the denoising nature of diffusion models, guiding generation in later denoising timesteps using edit instructions while preserving the input image layout. Since then, various works improved upon this by embedding the input image into the model's latent space~\cite{song2020denoising,kawar2023imagic,mokady2023null,parmar2023zero} or using cross-attention and self-attention mechanisms for realistic and targeted edits~\cite{hertz2022prompt,chefer2023attend,ge2023expressive,patashnik2023localizing,cao2023masactrl}. Recently, several methods train conditional diffusion models to follow user edit 
 instructions or spatial controls~\cite{zhang2023adding,brooks2023instructpix2pix}. 
However, these methods primarily focus on appearance editing, while our work enables both viewpoint and appearance control.

\myparagraph{Model customization.}%
While pre-trained models can generate common objects, users often wish to synthesize images with concepts from their own lives. This has given rise to the emerging technique %
of model personalization or customization~\cite{ruiz2022dreambooth,gal2022image,kumari2023multi}. These methods aim at embedding a new concept, e.g., pet dog, personal car, person, etc., into the output space of text-to-image models. This enables generating new images of the concept in unseen scenarios using the text prompt, e.g., my car in a field of sunflowers. To achieve this, various works fine-tune a small subset of model parameters~\cite{kumari2023multi,han2023svdiff,hu2021lora,tewel2023key} and/or optimize text token  embeddings~\cite{gal2022image,voynov2023p+,zhang2023prospect,alaluf2023neural} on the few images of the new concept with different regularizations~\cite{ruiz2022dreambooth,kumari2023multi}. More recently, encoder-based methods have been proposed that train a model on a vast dataset of concepts~\cite{shi2023instantbooth,arar2023domain,gal2023encoder,wei2023elite,li2023blip,valevski2023face0,ruiz2023hyperdreambooth,ye2023ip}, enabling faster customization. However, to our knowledge, no existing work allows for controlling the viewpoint in model customization. Given the ease of capturing multi-view images of a new concept, this work explores augmenting model customization with additional object viewpoint control.

\begin{figure*}[t]
    \centering
    \includegraphics[width=\linewidth]{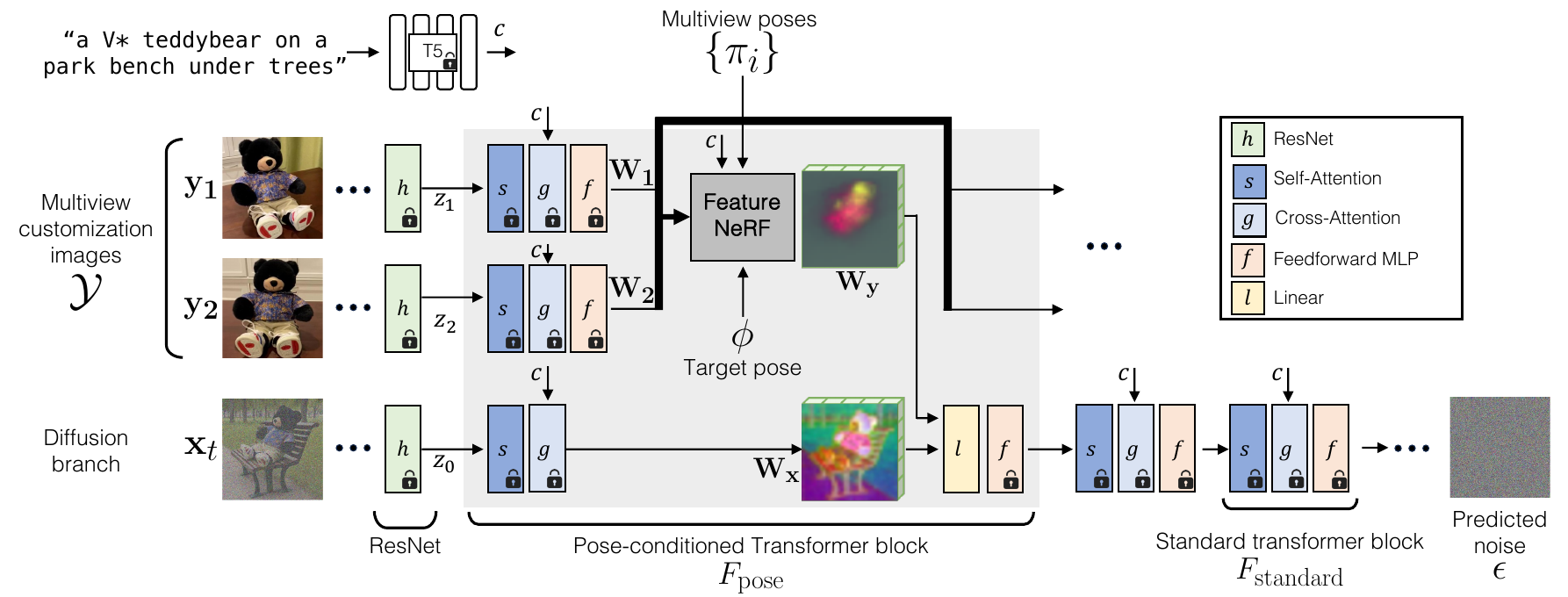}
    \vspace{-20pt}
    \caption{{\textbf{Overview.} We propose a model customization method that utilizes $N$ reference images defining the 3D structure of an object $\Y$ (we illustrate with $N=2$ views for simplicity). We modify the diffusion model U-Net with pose-conditioned transformer blocks. Our \textbf{Pose-conditioned transformer block} features a  FeatureNeRF module, which aggregates features from the individual viewpoints to target viewpoint $\phi$, as shown in detail in \reffig{feature_nerf}. The rendered feature $W_{y}$ is concatenated with the target noisy feature $W_{\mathbf{x}}$ and projected to the original channel dimension. We use the diffusion U-Net itself to extract features of reference images, as shown in the top row. We only fine-tune the new parameters in linear projection layer $l$ and FeatureNerF in $F_{\text{pose}}$ blocks. 
    }}
    \lblfig{method}
    \vspace{-10pt}
\end{figure*}

\myparagraph{View synthesis.} 
Novel view synthesis aims to render a scene from unseen camera poses, given multi-view images. Recently, the success of volumetric rendering-based approaches like NeRF~\cite{mildenhall2021nerf} have led to numerous follow-up works with better quality~\cite{barron2021mip,barron2023zip}, faster speed~\cite{muller2022instant,chen2022tensorf}, and fewer training views~\cite{yu2021pixelnerf,Niemeyer2021Regnerf,deng2022depth,tancik2021learned}. Recent works learn generative models with large-scale multi-view data to learn generalizable representations for novel view synthesis~\cite{liu2023zero,sargent2023zeronvs,zhou2023sparsefusion,chan2023genvs,wu2023reconfusion,liu2023syncdreamer,burgess2024viewpointtextualinversiondiscovering}. While our work draws motivation from this line of research, our goal differs - we aim to enable object viewpoint control in text-to-image personalization, rather than capturing novel views of real scenes. Concurrent to our work, ReconFusion~\cite{wu2023reconfusion} also trains a PixelNeRF~\cite{yu2021pixelnerf} in the latent space of latent diffusion models for 3D reconstruction. Different from this, we learn volumetric features in the intermediate attention layers. We also focus on model customization rather than scene reconstruction. 
Recently, Cheng \etal~\cite{cheng2024learning} and H\"ollein \etal ~\cite{hollein2024viewdiff} propose adding camera pose conditioning in text-to-image diffusion models while we focus on model customization. CustomNet~\cite{yuan2024customnet}, a concurrent work, also proposes to generate custom objects in a target viewpoint in a zero-shot manner. However, it focuses primarily on generating the new object in different backgrounds, whereas our method allows any new text prompt and viewpoint combination as a condition during inference.

\myparagraph{3D editing.}
Loosely related to our work, many works have been proposed for inserting and manipulating 3D objects within 2D real photographs by editing the image, using classic geometry-based approaches~\cite{karsch2011rendering,chen20133,kholgade20143d} or generative modeling techniques~\cite{yao20183d,zhang2020image,michel2023object,discoscene_2023, yenphraphai2024image}. Instead of editing a single image, our work aims to ``edit'' the model weights of a pre-trained diffusion model. Another relevant line of work edits~\cite{haque2023instruct,dong2024vica} or generates~\cite{raj2023dreambooth3d,tang2023dreamgaussian,shi2023mvdream, xu2023dmv3d,metzer2023latent} a 3D scene given a text prompt or image. 
Unlike these methods, we do not aim to edit/generate a multi-view consistent scene. Our goal is to provide additional viewpoint control when customizing text-to-image models. This enables specifying the object viewpoint while generating new backgrounds or composing multiple objects. Additionally, we show that our method achieves greater photorealism compared to a 3D editing method for this task.

\vspace{-2pt}
\section{Method}\lblsec{method}

Given multi-view images of a custom object, we aim to embed it in the text-to-image diffusion model. We construct our method in order to allow the generation of new variations of the object through text prompts while providing control of the object viewpoint. Our approach involves fine-tuning the pre-trained model while conditioning it on a 3D representation of the object learned in the diffusion model's feature space. In this section, we briefly overview the diffusion model and then explain our method in detail. 

\vspace{-2pt}
\subsection{Diffusion Models}

Diffusion models~\cite{sohl2015deep,ho2020denoising} are a class of generative models that sample images by iterative denoising of a random Gaussian distribution. The training of the diffusion model consists of a forward Markov process, where real data $\x_0$ is gradually transformed to random noise $\x_T \sim \mathcal{N} (\mathbf{0}, \mathbf{I})$ by sequentially adding Gaussian perturbations in $T$ timesteps, i.e., $\x_t = \sqrt{\alpha_t}\x_{0} + \sqrt{1 - \alpha_t}\epsilon$. The model is trained to learn the backward process, i.e., 

\begin{equation}
    \begin{aligned}
    p_\theta(\x_0 | \mathbf{c}) %
    = \int \Bigr [ p_{\theta} (\x_{T}) \prod p_{\theta}^t(\x_{t-1} | \x_t, \mathbf{c}) \Bigr ] d\x_{1:T}, 
    \end{aligned}\label{eq:diffusionformulation}
\end{equation}
The training objective maximizes the variational lower bound, which can be simplified to a simple reconstruction loss: 
\begin{equation}
    \begin{aligned}
     \mathbb{E}_{\x_t,t,\mathbf{c}, \epsilon \sim \mathcal{N} (\mathbf{0}, \mathbf{I})} [w_t||\epsilon - \epsilon_{\theta} (\x_t, t, \mathbf{c}) ||],
    \end{aligned}\label{eq:diffusion}
\end{equation}
where $\mathbf{c}$ can be any modality to condition the generation process. The model is trained to predict the noise added to create the input noisy image $\x_t$. During inference, we gradually denoise a random Gaussian noise over a fixed number of timesteps. Various proposed sampling strategies~\cite{song2020denoising,lu2022dpm,karras2022elucidating} reduce the number of sampling steps compared to the typical $1000$ timesteps in training.  In our work, we use the Stable Diffusion-XL (SDXL)~\cite{podell2023sdxl} as the pre-trained text-to-image diffusion model. It is based on the Latent Diffusion Model (LDM)~\cite{rombach2022high}, which is trained in an autoencoder~\cite{kingma2013auto} latent space.

\subsection{Customization with Object Viewpoint Control}
Model customization aims to condition the model on a new object, given $N$ images of the object $\Y=\{\y_i\}_{i=1}^N$, i.e., to model $p(\x | \Y, \mathbf{c} )$ with text prompt $\mathbf{c}$. In contrast, we additionally condition the model on the object viewpoint, allowing more control in the generation process. Thus, given a set of multi-view images $\{\y_i\}_{i=1}^N$ and the corresponding camera poses $ \{\pi_i\}_{i=1}^N$, 
our goal is to learn the conditional distribution $p(\x | \{(\y_i, \pi_i)\}_{i=1}^N, \mathbf{c}, \phi)$, where $\mathbf{c}$ is text prompt and $\phi$ is the camera pose corresponding to the target viewpoint. To achieve this, we fine-tune a pre-trained text-to-image diffusion model, which models $p(\x | \mathbf{c})$, with the additional conditioning of camera pose $\phi$ given posed reference images $\{\y_i, \pi_i \}_{i=1}^N$. 

\myparagraph{Model architecture.} In \reffig{method}, we show the overall architecture, with an emphasis on our added pose-conditioning. Each block in the diffusion model U-Net~\cite{ronneberger2015u} consists of a ResNet~\cite{resnet}, denoted as $h$, followed by several transformer layers~\cite{vaswani2017attention}. Given the output of an intermediate ResNet layer $\z$, a standard transformer layer, $F_{\text{standard}}(\z, \mathbf{c})$, consists of a self-attention layer, denoted as $s$, followed by cross-attention with the text prompt, denoted as $g$, and a feed-forward MLP, denoted as $f$. We modify a subset of these transformer layers to incorporate pose conditioning as we explain next.

\myparagraph{Pose-conditioned transformer layer.} 
We denote the pose-\\conditioned transformer layer as $F_{\text{pose}}(\z_0, \{\z_i, \pi_i\})$, where $\z_0$ is the intermediate target feature (diffusion branch in \reffig{method}) and $\{\z_i\}$ are the input features corresponding to multi-view reference images (top two rows in \reffig{method}).
We extract spatial features $\{\W_i\}$ from $\{\z_i\}$ using components of pre-trained U-Net itself, i.e., $F_{\text{standard}}(\z_i, \c)$. To condition the diffusion branch on $\phi$, we learn a radiance field, denoted as FeatureNeRF, from $\{\W_i, \pi_i \}$ in a feed-forward manner~\cite{yu2021pixelnerf}. The predicted FeatureNeRF is then rendered from the target viewpoint $\phi$ to obtain view-dependent feature map $\W_{y}$.

In the main diffusion branch, we extract the intermediate feature map after the self and cross-attention layers, i.e., $\W_{\x} = g(s(\z_0), \mathbf{c})$. We concatenate $\W_{\x}$ with the rendered features $\W_{y}$ and then project it into the original feature dimension using a linear layer. Thus, the pose conditioned transformer layer, $F_{\text{pose}}(\z_0, \{\z_i, \pi_i\}, \mathbf{c}, \phi)$ performs:

\begin{figure}[!t]
    \includegraphics[width=\linewidth]{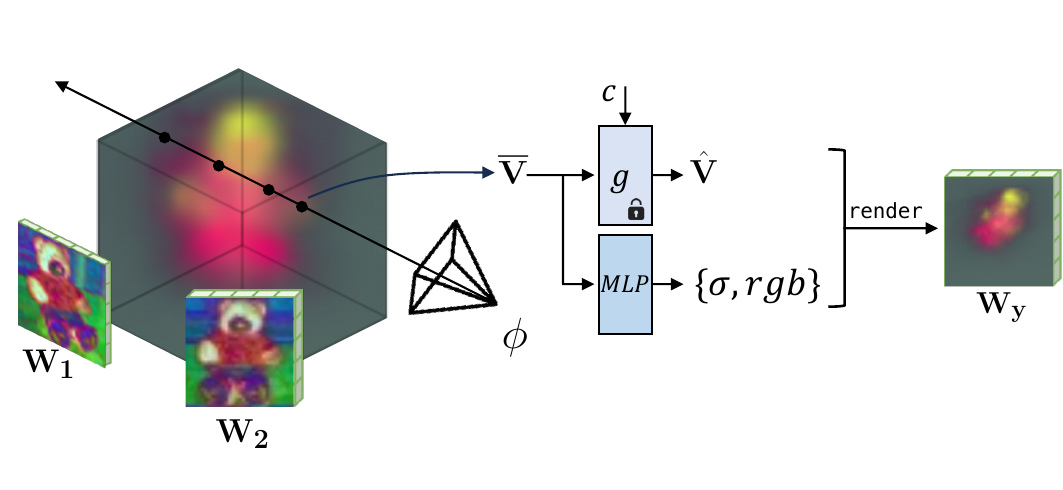}
    \vspace{-20pt}
    \caption{\textbf{FeatureNeRF}. We predict volumetric features $\overline{\mathbf{V}}$ for each 3D point in the grid using reference features $\{\W_i\}$ (\refeq{featurenerf1}). Given this feature, we predict the density $\sigma$ and color $rgb$ using a 2-layer MLP and use the predicted density $\sigma$ to render $\hat{\mathbf{V}}$ (which has been updated with text cross-attention $g$). The $rgb$ is only used to calculate reconstruction loss during training. 
    }
    \vspace{-10pt}
\label{fig:feature_nerf}
\end{figure}

\vspace{-3mm}
\begin{equation}
\begin{aligned}
\W_{i} = F_{\text{standard}}(\z_i, &\mathbf{c}), \hspace{2mm} \W_{y} = \text{FeatureNeRF}(\{\W_i, \pi_i\}, \mathbf{c}, \phi) \\ 
&F_{\text{pose}} = f(l(\W_y \oplus \W_{\x}))
\end{aligned}
\end{equation}

\noindent where $l$ is a learnable weight matrix, which projects the feature into the input space of feed-forward layer $f$. We initialize $l$ such that the contribution from $\W_{y}$ is zero at the start of training.

\myparagraph{FeatureNeRF.} Here, we describe the aggregation of individual features $\W_i$ with poses $\pi_i$ into a feature map $\W_{y}$ from pose $\phi$. Given a target ray with direction $\mathbf{d}$ from target viewpoint $\phi$, we sample points $\mathbf{p}$ along the ray and project it to the image plane of each given view $\pi_i$. The  projected coordinate is denoted as $\pi_i^{\mathbf{p}}$. We then sample the feature from this coordinate in $\W_i$, predict a feature for the 3D point $\mathbf{p}$, and aggregate the $N$ predicted features from each view with function $\psi$:

\begin{equation}
    \begin{aligned}
        \mathbf{V}_i^{\mathbf{p}} = &\text{MLP}(\text{Sample}(\W_i; \pi_i^{\mathbf{p}}), \gamma (\mathbf{d}), \gamma(\mathbf{p})), \; {i=1, ..., N} \\
        &\overline{\mathbf{V}}^{\mathbf{p}} = \psi(\mathbf{V}_1^{\mathbf{p}}, ..., \mathbf{V}_N^{\mathbf{p}}),    
    \end{aligned}\label{eq:featurenerf1}
\end{equation}

\noindent where $\gamma$ is the frequency encoding. We use the weighted average~\cite{reizenstein21co3d} as the aggregation function $\psi$, where a linear layer predicts the weights based on $\mathbf{V}_i$, $\pi_i$, and $\phi$. For each reference view, $\mathbf{d}$ and $\mathbf{p}$ are first transformed in the view coordinate space~\cite{yu2021pixelnerf}. Given the feature  $\overline{\mathbf{V}}$ (superscript ${\mathbf{p}}$ is dropped for simplicity) for the 3D point, we predict the density and color using a linear layer:
\begin{equation}
    \begin{aligned}
     (\sigma, \mathbf{C}) = \text{MLP}(\overline{\mathbf{V}}),
    \end{aligned}\label{eq:featurenerf2}
\end{equation}
and also update the aggregated feature with text prompt $\mathbf{c}$ using cross-attention:
\begin{equation}
    \begin{aligned}
     \hat{\mathbf{V}} = \text{CrossAttn}(\overline{\mathbf{V}}, \mathbf{c}).
    \end{aligned}\label{eq:featurenerf_update}
\end{equation}
\noindent We then render this updated feature volume using the predicted densities:
\begin{equation}
    \begin{aligned}
     & \W_{y}(r) = \sum_{j=1}^{N_f} T_j(1 - \exp(-\sigma_j\delta_j)) \hat{\mathbf{V}}_j,
    \end{aligned}\label{eq:predicted_feat}
\end{equation}
where $r$ is the target ray, $\hat{\mathbf{V}}_j$ is the feature corresponding to the $j^{th}$ point along the ray, $\sigma_j$ is the predicted density of that point, $N_f$ is the number of sampled points along the ray between the near and far plane of the camera, and $T_j =\exp(-\sum_{k=1}^{j-1}\sigma_k\delta_k)$ handles occlusion until that point. 

We build our FeatureNeRF design based on PixelNeRF~\cite{yu2021pixelnerf} but update the aggregated features with text cross-attention and use learnable weighted averaging to aggregate reference view features. Through this layer, our focus is on learning 3D features that the 2D diffusion model can use rather than learning NeRF in a feature space~\cite{ye2023featurenerf,kerr2023lerf}.

\myparagraph{Training loss.} Our training objective includes learning 3D consistent FeatureNeRF modules, which can contribute to the final goal of reconstructing the target concept in diffusion models output space. Thus, we fine-tune the model using the sum of training losses corresponding to FeatureNeRF and the default diffusion model reconstruction loss:
\begin{equation}
    \begin{aligned}
     & \mathcal{L}_{\text{diffusion}} = \sum_r M w_t||\epsilon - \epsilon_{\theta} (\x_t, t, \mathbf{c}) ||, 
    \end{aligned}\label{eq:loss_diffusion_masked}
\end{equation}
where $M$ is the object mask, with the reconstruction loss being calculated only in the object mask region. The losses corresponding to FeatureNeRF consist of RGB reconstruction loss:
\begin{equation}
    \begin{aligned}
    &\mathcal{L}_{\text{rgb}} = \sum_{r}|| M(r) (\mathbf{C}_{gt}(r) - \sum_{j=1}^{N_f} T_j(1 - \exp(-\sigma_j\delta_j))\mathbf{C})  ||,
    \end{aligned}\label{eq:loss_rgb}
\end{equation}

\noindent and two mask-based losses as we only wish to model the object -- (1) silhouette loss~\cite{ravi2020accelerating} $\mathcal{L}_{\text{s}}$ which forces the rendered opacity to be similar to object mask, and (2) background suppression loss~\cite{boss2021nerd,boss2022samurai} $\mathcal{L}_{\text{bg}}$ which enforces the density of all background rays to be zero. 
\begin{equation}
    \begin{aligned}
     \mathcal{L}_{\text{s}} &= \sum_{r}|| M(r) - \sum_{j=1}^{N_f} T_j(1 - \exp(-\sigma_j\delta_j))  || \\
     &\mathcal{L}_{\text{bg}} =  \sum_{r} (1-M(r)) \sum_{j=1}^{N_f}||(1 - \exp(-\sigma_j\delta_j)) ||,
    \end{aligned}\label{eq:loss_background}
\end{equation} %

\noindent Thus, the final training loss is: 
\vspace{-1mm}
\begin{equation}
    \begin{aligned}
    &\mathcal{L} = \mathcal{L}_{\text{diffusion}} + \lambda_{\text{rgb}} \mathcal{L}_{\text{rgb}} + \lambda_{\text{bg}}\mathcal{L}_{\text{bg}} + \lambda_{\text{s}}\mathcal{L}_{\text{s}},
    \end{aligned}\label{eq:loss}
\end{equation}
where $M$ is the object mask and $\lambda_{\text{rgb}}$, $\lambda_{\text{bg}}$, and $\lambda_{\text{s}}$ are hyperparameters to control the rendering quality of intermediate images vs. the final denoised image and are fixed across all experiments. 
We assume access to the object's mask in the image, which is used to calculate these losses. The losses for FeatureNeRF are averaged across all pose-conditioned transformer layers.

\begin{figure*}[!t]
    \centering
    \includegraphics[width=\linewidth]{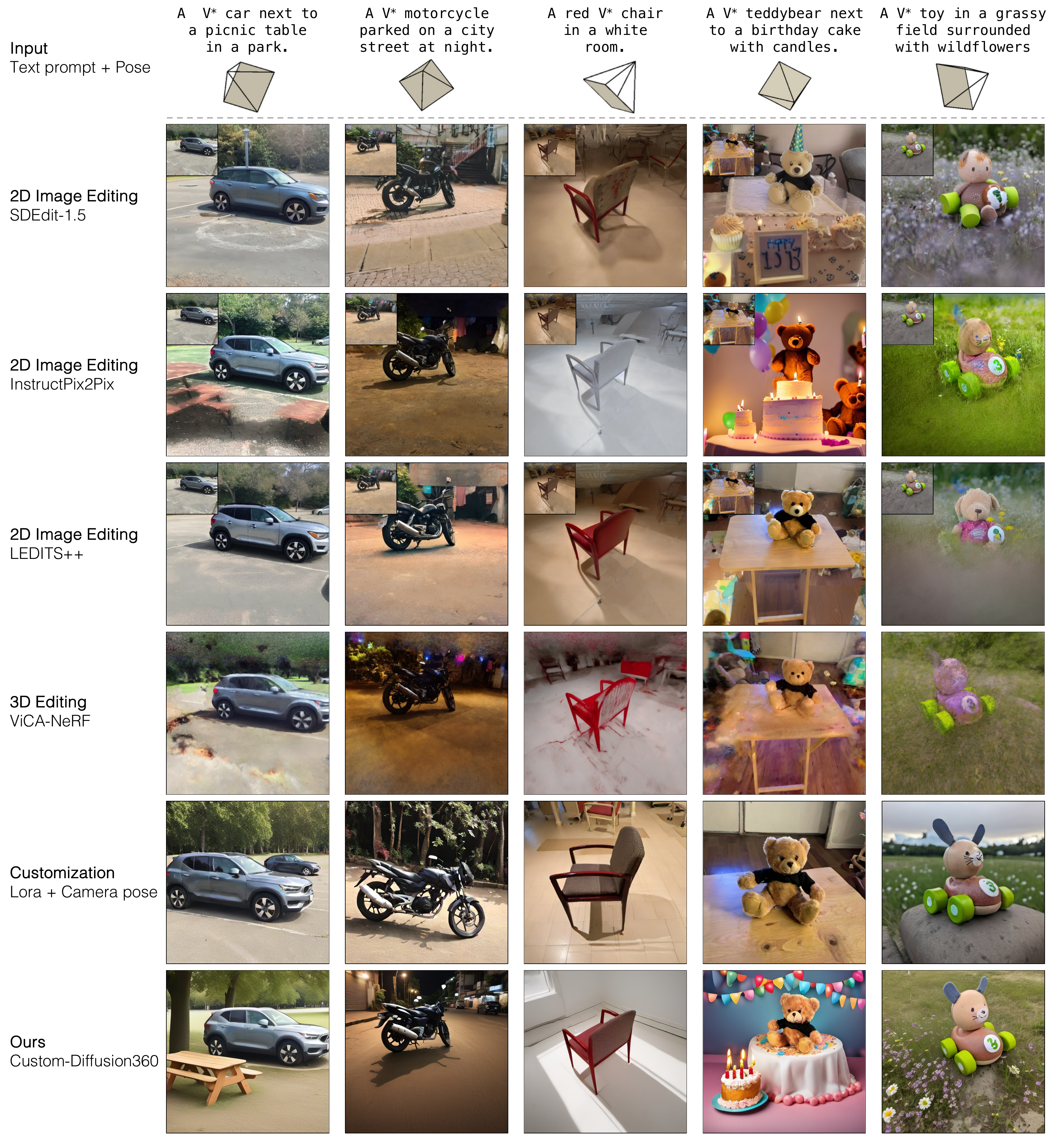}
    \vspace{-12pt}
    \caption{{\textbf{Qualitative comparison.} Given a particular target pose, we show the qualitative comparison of our method with (1) Image editing methods \textit{SDEdit}, \textit{InstructPix2Pix}, and \textit{LEDITS++}, which edit a NeRF-rendered image from the input pose, (2) \textit{ViCA-NeRF}, a 3D editing method that trains a NeRF model for each input prompt, and (3) \textit{LoRA + Camera pose}, our proposed baseline where we concatenate camera pose information to text embeddings during LoRA fine-tuning. Our method performs on par or better in keeping the target identity and poses while incorporating the new text prompt---e.g., putting a picnic table next to the SUV car ($1^{\text{st}}$ column)---and following multiple text conditions---e.g., turning the chair red and placing it in a white room ($3^{\text{rd}}$ column). {\menlo V$^*$} token is used only in ours and the LoRA + Camera pose method. Ground truth rendering from the given pose is shown as an inset in the first three rows. We show more sample comparisons in \reffig{result_appendix} of Appendix.  
    }}
    \lblfig{result_qualitative}
    \vspace{-10pt}
\end{figure*}

\begin{figure*}[!t]
    \centering
    \includegraphics[width=\linewidth]{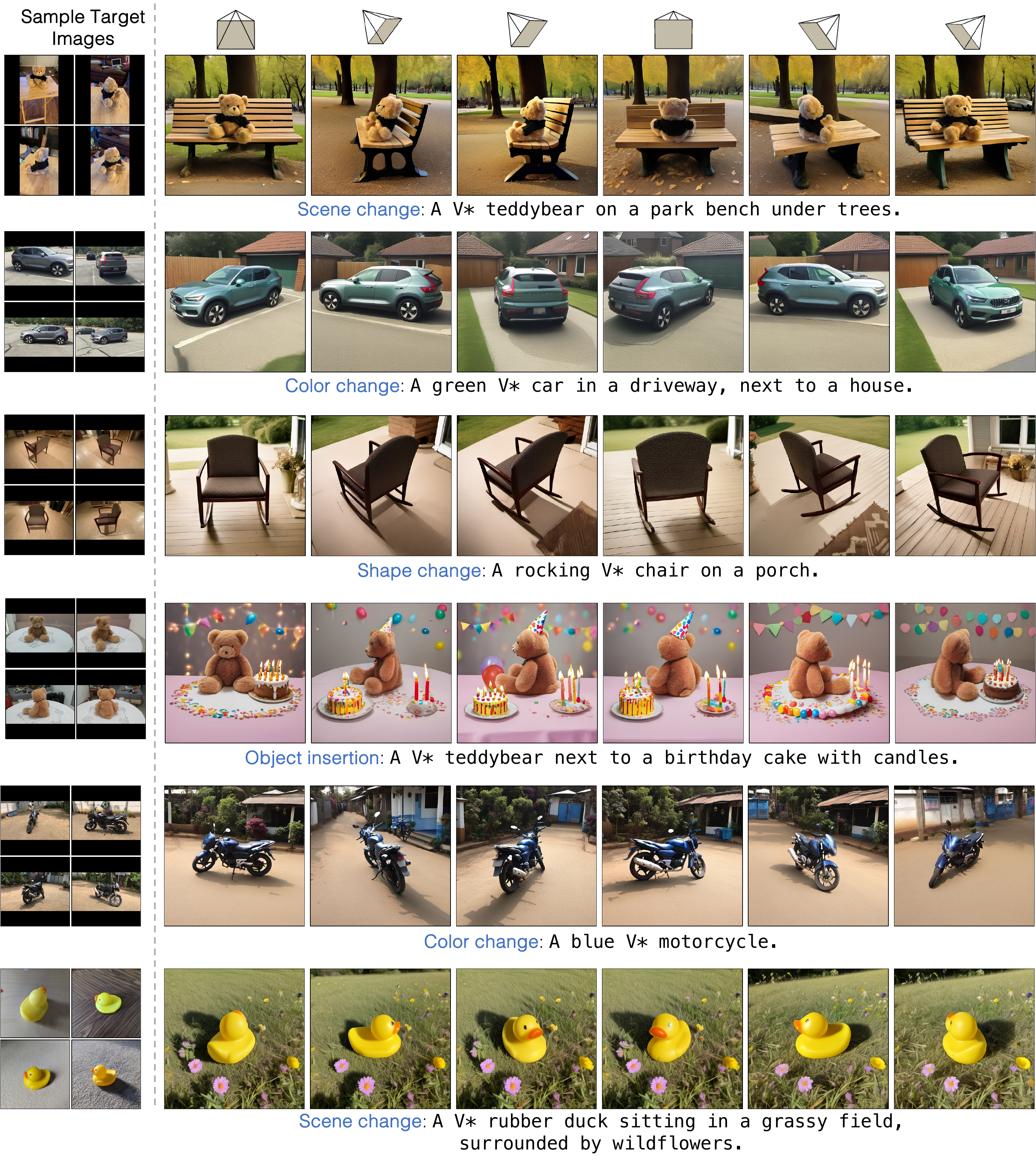}
    \caption{\textbf{Qualitative samples with varying object viewpoint and text prompt}. Our method learns the identity of custom objects while allowing the user to control the object viewpoint and generating the object in new contexts using the text prompt, e.g., changing the background scene or object color and shape. In each row, the images were generated with the same seed while changing the object viewpoint in a turntable manner. Note that each image in a row is independently generated.    
    \reffig{result2_appendix} in the Appendix shows more such samples. 
    }
    \vspace{10pt}
\label{fig:more_results1}
\end{figure*}

\myparagraph{Inference.} During inference, to balance the text vs. reference view conditions in the final generated image, we combine text and image guidance~\cite{brooks2023instructpix2pix} as shown below: 
\vspace{-1mm}
\begin{equation}
    \begin{aligned}
     \hat{\epsilon}_{\theta}(\x_t, I = \{\y_i, \pi_i \}_{i=1}^N, &\mathbf{c}) = \epsilon_{\theta}(\x_t, \varnothing, \varnothing) \\
     & + \lambda_{I}(\epsilon_{\theta}(\x_t, I, \varnothing)  - \epsilon_{\theta}(\x_t, \varnothing, \varnothing) ) \\
     & + \lambda_{c}(\epsilon_{\theta}(\x_t, I, \mathbf{c})  - \epsilon_{\theta}(\x_t, I, \varnothing) ),
    \end{aligned}\label{eq:inference_ours}
\end{equation}
where $\lambda_I$ is the image guidance scale and $\lambda_c$ is the text guidance scale. Increasing the image guidance scale increases the generated image's similarity to the reference images. Increasing the text guidance scale increases the generated image's consistency with the text prompt. 

\myparagraph{Training details.}
During training, we sample the $N$ views equidistant from each other and use the first as the target viewpoint and the others as references. We modify $12$ transformer layers with pose conditioning out of $70$ transformer layers in Stable Diffusion-XL. For rendering, we sample $24$ points along the ray. The new concept is described as ``V$^*$ {category}'', with V$^*$ as a trainable token embedding~\cite{kumari2023multi,gal2022image}. Furthermore, to reduce overfitting~\cite{ruiz2022dreambooth}, we use generated images of the same category, such as random car images with ChatGPT-generated captions~\cite{chatgpt}. These images are randomly sampled $25\%$ of the time during training. We also drop the text prompt with $10\%$ probability to be able to use classifier-free guidance. We provide more implementation details in \refapp{details}.

\section{Experiments}

\myparagraph{Dataset.} 
We select 14 custom objects from the CO3Dv2~\cite{reizenstein21co3d} and NAVI~\cite{jampani2024navi} datasets. Specifically, we select $4$ categories with $3$ instances each from the CO3Dv2 dataset -- car, chair, teddy bear, and motorcycle. From NAVI, we select $2$ unique, toy-like objects. A representative image of each concept is shown in the supplemental material. We use the camera poses provided in the dataset. For each instance, we sample $\sim 100$ images, using half for training and half for evaluation. The camera poses are normalized such that the mean of camera location is the origin, and the first camera is at unit norm~\cite{zhang2022relpose}.

\begin{table}[!t]
\centering
\setlength{\tabcolsep}{5pt}
\resizebox{\linewidth}{!}{
\begin{tabular}{@{\extracolsep{4pt}}l  cc c }
\toprule

\textbf{Method}   & Text Alignment & Image  Alignment  & Photorealism \\
\midrule
SDEdit & 40.06 $\pm$ 2.68$\%$ & 36.08 $\pm$ 2.80$\%$ & 33.11  $\pm$ 2.82$\%$ \\
vs. Ours &  \textbf{59.40} $\pm$ 2.68$\%$ & \textbf{63.92} $\pm$ 2.80$\%$  & \textbf{66.89}  $\pm$  3.18$\%$ \\
\cdashline{1-4}
InstructPix2Pix & 44.79 $\pm$ 2.58$\%$ & 29.34  $\pm$     2.24 $\%$ & 27.61  $\pm$     2.63$\%$ \\
vs. Ours & \textbf{55.21} $\pm$ 2.58$\%$  & \textbf{70.66} $\pm$     2.24 $\%$ & \textbf{72.39} $\pm$     2.63$\%$  \\
\cdashline{1-4}
LEDITS++ & 32.47 $\pm$ 2.39$\%$ & 35.86 $\pm$ 2.50$\%$ & 26.18  $\pm$ 2.82$\%$  \\
vs. Ours  &  \textbf{67.53}$\pm$ 2.39\%  & \textbf{64.14} $\pm$ 2.50\%  & \textbf{73.82}  $\pm$ 2.82$\%$  \\
\cdashline{1-4}
Vica-NeRF & 27.13 $\pm$ 2.83$\%$& 24.36 $\pm$ 3.35$\%$ & 12.90  $\pm$ 2.67$\%$\\
vs. Ours & \textbf{72.87} $\pm$ 2.83$\%$ & \textbf{75.64} $\pm$ 3.35 $\%$ &  \textbf{87.10} $\pm$  2.67 $\%$ \\
\cdashline{1-4}
LoRA + Camera pose & 32.26 $\pm$ 2.67$\%$& \textbf{66.97} $\pm$ 2.50 $\%$ & \textbf{52.51}  $\pm$ 2.75$\%$\\
vs. Ours & \textbf{67.64} $\pm$ 2.67$\%$ & 33.03 $\pm$ 2.50$\%$ & 47.49   $\pm$ 2.75$\%$\\

\bottomrule
\vspace{-20pt}
\end{tabular}
}
\caption{\textbf{Human preference evaluation}. Our method is preferred over almost all baselines for text alignment, image alignment to the target concept, and photorealism. We find that LoRA + Camera pose overfits the training images, as shown in \reffig{result_qualitative}. }
\vspace{-2pt}
\label{tbl:human_eval}
\end{table}

\begin{table}[!t]
\centering
\setlength{\tabcolsep}{5pt}
\resizebox{\linewidth}{!}{
\begin{tabular}{l  c c}
\toprule
\textbf{Method}
& \shortstack[c]{\textbf{Angular} \textbf{error}}
& \multicolumn{1}{@{} c}{\shortstack[c]{\textbf{Camera} \textbf{center error}}}  \\
\midrule
\textbf{Ours}  &   \textbf{14.19} &  \textbf{0.080} \\
\textbf{LoRA + Camera pose}  &  41.14	&  0.305  \\
\bottomrule
\vspace{-2pt}
\end{tabular}
}
\vspace{-14pt}
\caption{\textbf{Accuracy of object viewpoint condition} in generated images by ours and the LoRA + Camera pose baseline method. We observe that the baseline usually overfits the training images and does not respect the target viewpoint condition with new text prompts.
}

\label{tbl:camera_pose}
\vspace{-8pt}
\end{table}

\myparagraph{Baselines.}
We compare with three types of relevant baselines -- (1) First, we compare against 2D image editing using
$3$ recent, publicly available methods -- LEDITS++~\cite{brack2023ledits++}, InstructPix2Pix~\cite{brooks2023instructpix2pix}, and SDEdit~\cite{meng2021sdedit} with Stable Diffusion-1.5 (and SDXL in \refapp{results1}). As image editing methods do not inherently support viewpoint manipulation, we first render a NeRF~\cite{tancik2023nerfstudio} of the input scene with the target viewpoint and then edit the rendered image. (2) Secondly, we use a customization-based method, LoRA+Camera pose, where we modify LoRA~\cite{loraimplementation,hu2021lora} by concatenating the camera pose to the text embeddings, following recent work Zero-1-to-3~\cite{liu2023zero}. (3) Lastly, we test ViCA-NeRF~\cite{dong2024vica}, a 3D editing method that trains a NeRF for each new text prompt. In \refapp{details}, we provide more details on implementation and hyperparameters for each baseline.

\myparagraph{Evaluation metrics.}
To create an evaluation set, we generate $16$ prompts per object category using ChatGPT~\cite{chatgpt}. We instruct ChatGPT to propose four types of prompts: scene change, color change, object composition, and shape change. 
We then manually inspect them to remove implausible or overly complicated text prompts~\cite{wang2023evaluating}. \reftbl{prompts_eval} in \refapp{eval_details} lists all the evaluation prompts. We evaluate (1) the customization quality of the generated image and (2) its adherence to the specific pose.

First, to measure customization quality, we use a pairwise human preference study. A successful customization is comprised of several aspects: alignment to the target concept, alignment to the input text prompt, and photorealism of the generated images. In total, we collect $\sim1000$ responses per pairwise study against each baseline using Amazon Mechanical Turk. We also evaluate our method and baselines on other standard metrics like CLIP Score~\cite{radford2021learning} and DINOv2~\cite{oquab2023dinov2} image similarity~\cite{ruiz2022dreambooth} to measure the text and image alignment. 

To measure whether the generated custom object adheres to the specified viewpoint, we use a pre-trained model, RayDiffusion~\cite{zhang2023cameras}, to predict the pose from the generated images and calculate its error relative to the input camera pose. More details about evaluation are provided in \refapp{eval_details}.

\begin{figure}[!t]
    \includegraphics[width=\linewidth]{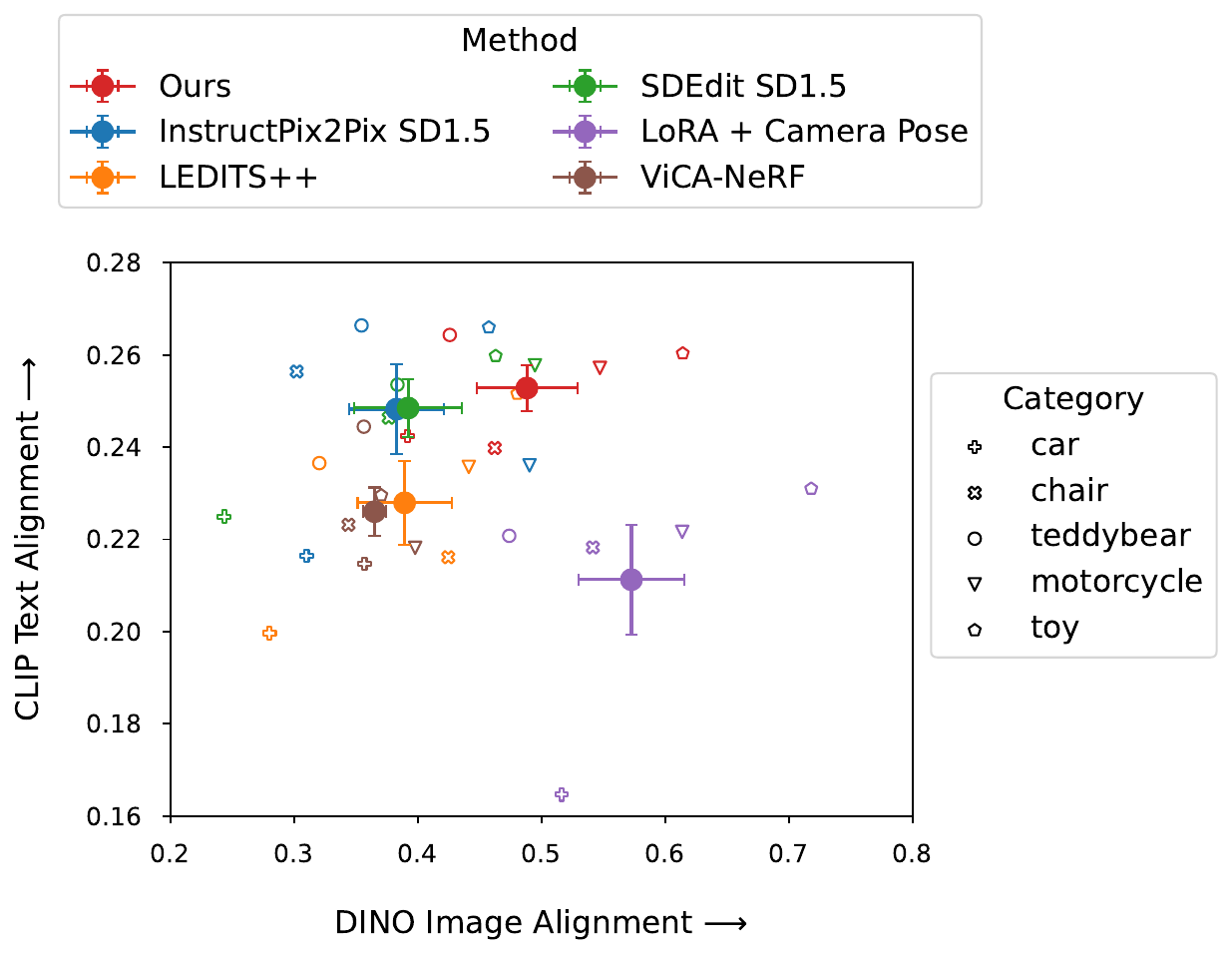}
    \vspace{-20pt}
    \caption{\textbf{Quantitative comparison}. We show CLIP scores (higher is better) vs. DINO-v2 scores (higher is better). We plot the performance of each method on each category and the overall mean and standard error (highlighted). Our method results in higher CLIP text alignment while maintaining visual similarity to target concepts, as indicated by DINO-v2 scores. The text alignment of our method compared to SDEdit and InstructPix2Pix is only marginally better as these methods incorporate the text prompt but at the cost of photorealism, as we show in \reftbl{human_eval}.
    }
    \vspace{-10pt}
\label{fig:dino_clip_all_methods}
\end{figure}

\subsection{Results}\lblsec{results1}

\begin{figure*}[!t]
    \includegraphics[width=\linewidth]{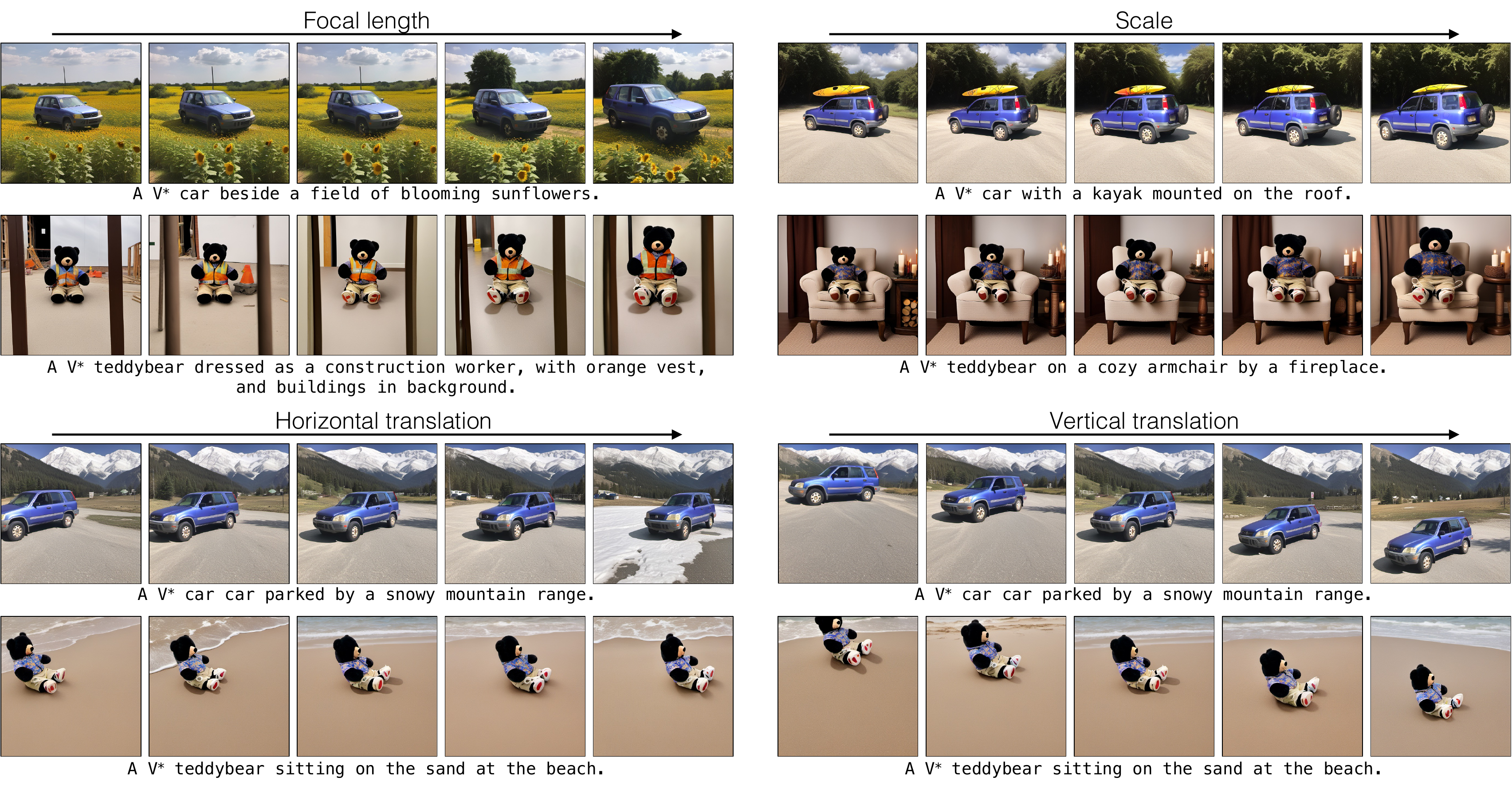}
    \vspace{-22pt}
    \caption{\textbf{Extrapolating object viewpoint from training viewpoints}. Our method can generalize to different viewpoints, including those not within the training distribution. \textit{Top left:} We vary the focal length from $\times 0.8$ to $\times 1.4$ of the original focal length. \textit{Top right:} We vary the camera position towards the image plane along the $z$ axis. \textit{Bottom row:} We vary the camera position along the horizontal and vertical axis. }
    \vspace{-8pt}
\label{fig:focal_change}
\end{figure*}

\myparagraph{Generation quality and adherence.} First, we measure the quality of the generation -- adherence to the text prompt, the identity preservation to the customized objects, and photorealism -- irrespective of the object viewpoint. Recall that for each concept, we curate 16 prompts. For each prompt, we generate $3$ images at each viewpoint, covering $6$ target viewpoints, resulting in $288$ images per concept. \reftbl{human_eval} shows the pairwise human preference for our method vs. baselines. Our method is preferred over all baselines except LoRA + Camera pose, which we observe to overfit on training images, thus producing higher image alignment. \reffig{dino_clip_all_methods} shows the CLIP vs. DINO scores for all methods. Ideally, a method should have both a high CLIP score and a DINO score, but often, there is a trade-off between text and image alignment. Our method has on-par or better text alignment relative to the baselines, while having better image alignment. We observe that image-editing baselines often require careful hyperparameter tuning for each image. We select the best-performing hyperparameters and keep them fixed across all experiments. The camera pose corresponding to the target object viewpoint is uniformly sampled from $\sim 50$ validation poses not used during training. We also randomly perturb the camera position or focal length. \reffig{views_vis} in \refapp{eval_details} shows sample training and perturbed validation camera poses for the car object.

\myparagraph{Accuracy of object viewpoint.}
Previously, we evaluated our method purely on image customization benchmarks. Next, we evaluate the accuracy of the object viewpoint conditioning. \reftbl{camera_pose} shows the mean angular error and camera center error between the generated object's pose, predicted using RayDiffusion~\cite{zhang2023cameras}, and the input pose. We only compare with LoRA + Camera pose, as only this baseline takes the camera pose for the target object viewpoint as input. We observe that it often overfits training images and fails to generate the object in the correct viewpoint with new text prompts. We evaluate this metric only on the objects from the CO3Dv2 dataset with validation camera poses, as RayDiffusion has been trained on CO3Dv2 and struggles with other unique objects. 

\myparagraph{Qualitative comparison.} 
We show the qualitative comparison of our method with the baselines in \reffig{result_qualitative}. 
We observe that image editing methods can fail to generate photorealistic results. In the case of LoRA + Camera pose, it fails to generalize and overfits to the training views ($5^{\text{th}}$ row). Finally, the 3D editing-based method ViCA-NeRF maintains 3D consistency but generates blurred images, especially for text prompts that change the background. \reffig{more_results1} shows more samples with different text prompts and object viewpoints for our method.

\begin{figure}[!t]
    \includegraphics[width=0.8\linewidth]{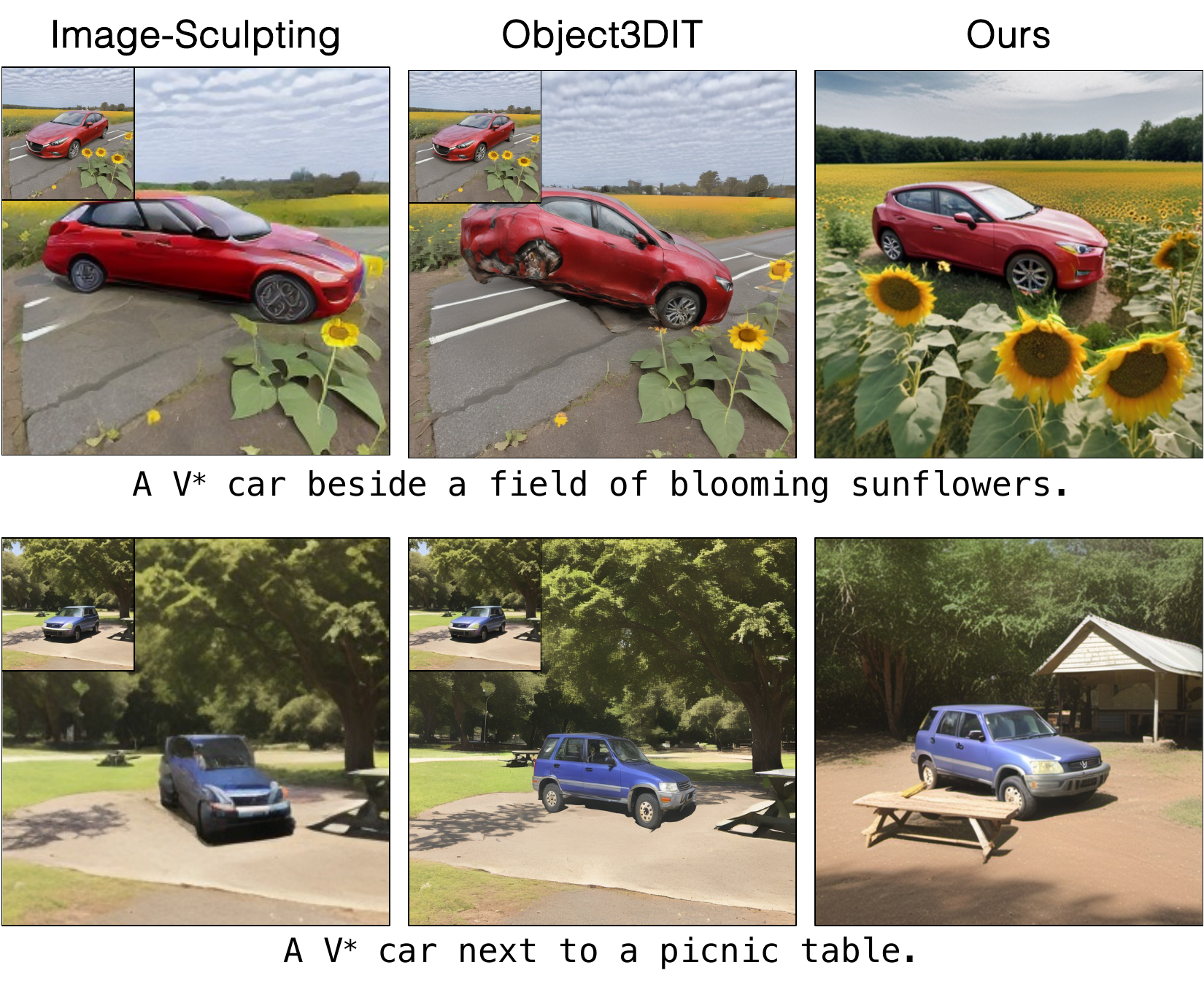}
    \caption{\textbf{Comparison to 3D-aware editing methods}. We first generate the image of the custom object using LoRA+DreamBooth (shown as an inset) and then use the 3D-aware editing method to edit and rotate the object to a target viewpoint. We show qualitative samples generated by our method ($3^{\text{rd}}$ column) with approximately the same target viewpoint as input. Object3DIT and Image-Sculpting lead to lower fidelity edits than images generated by our method ($3^{\text{rd}}$ column) with the target viewpoint directly as the input condition.}
\label{fig:3dediting}
\end{figure}

\myparagraph{Additional comparison to customization + 3D-aware image editing.} We further compare against a two-stage approach that first generates an image of the custom object using LoRA+DreamBooth \cite{loraimplementation,ruiz2022dreambooth} and then edits the object to a target viewpoint using two recent 3D-aware image editing methods, Image Sculpting~\cite{yenphraphai2024image} and Object3DIT~\cite{michel2023object}. For each prompt, we generate $3$ images, then edit and rotate the object to 6 different viewpoints. This results in 288 images per concept, similar to our evaluation setting. We compare against this on only the three car objects since Image-Sculpting uses Adobe Photoshop's generative fill~\cite{generative_fill} as one of the intermediate steps, which requires manually inpainting each image. The CLIP scores for Image Sculpting and Object3DIT are $0.26$ and $0.27$, respectively, compared to our score of $0.25$.
However, their DINO scores at $0.24$ and $0.40$ are substantially lower than our $0.48$. 
As shown in \reffig{3dediting}, both methods lead to lower-fidelity results. Object3DIT struggles in many scenarios due to its training on a synthetic dataset, and Image Sculpting's performance is highly dependent on single image-to-3D methods like Zero-1-to-3~\cite{liu2023zero} used in its pipeline.

\myparagraph{Generalization to novel viewpoints.} Since our method learns a 3D radiance field, we can also extrapolate to unseen object viewpoints at inference time as shown in \reffig{focal_change}. We generate images while varying the camera distance from the object (scale), focal length, or camera position along the horizontal and vertical axis.

\begin{figure}[!t]
    \includegraphics[width=\linewidth]{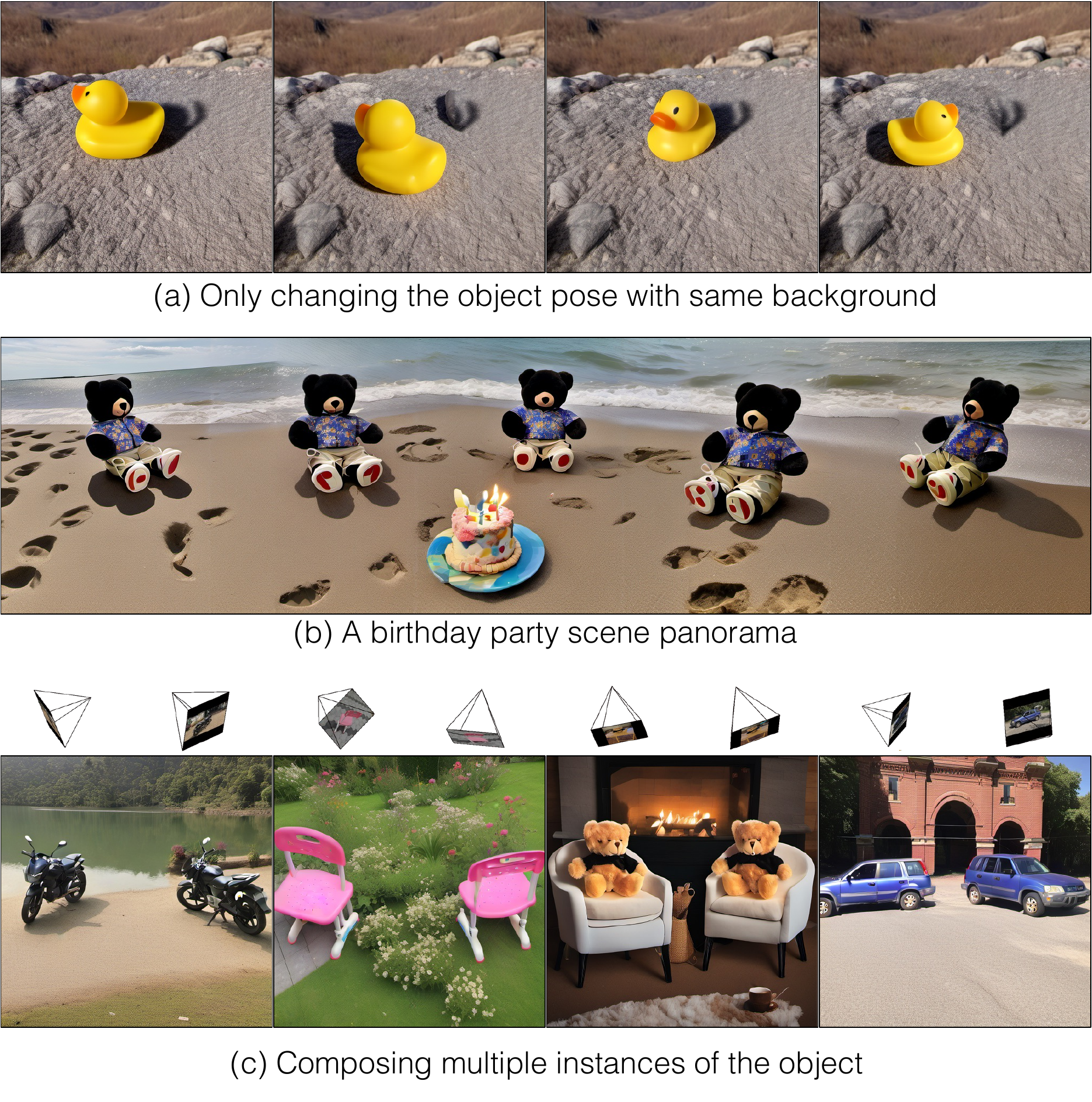}
    \vspace{-15pt}
    \caption{\textbf{Applications}. \textit{$1^{\text{st}}$ row}: Our method can be combined with other image editing methods as well. We use SDEdit with our method to in-paint the car and rubber duck from different viewpoints while keeping the same background. \textit{$2^{\text{nd}}$ row}: We can generate interesting panorama shots by controlling the object viewpoint independently in each grid. \textit{$3^{\text{rd}}$ row}: We can also compose the radiance field predicted by FeatureNeRF to control the relative pose while generating multiple instances of the object. }
\label{fig:application}
\end{figure}

\myparagraph{Applications.}
Our method can be combined with existing image editing methods as well. \reffig{application}a shows an example where we use SDEdit~\cite{meng2021sdedit} to in-paint the object with different viewpoints while keeping the same background. We can also generate interesting panoramas using MultiDiffusion~\cite{bar2023multidiffusion}, where the object viewpoint in each grid is controlled by our method, as shown in \reffig{application}b. Moreover, since we learn a 3D consistent FeatureNeRF for the new object, we can compose multiple instances of the object~\cite{song2023totalrecon}, with each instance in a different viewpoint. \reffig{application}c shows an example of two teddy bears facing each other and sitting on armchairs. Here, we additionally use DenseDiffusion~\cite{densediffusion} to modulate the attention maps and guide the generation of each object instance to only appear in the corresponding region predicted by FeatureNeRF. At the same time, the attention maps of the empty region predicted by FeatureNeRF are modulated to match the part of the text prompt describing the image's background.

\begin{table}[!t]
\centering
\setlength{\tabcolsep}{5pt}
\resizebox{\linewidth}{!}{
\begin{tabular}{l c c c c c c}
\toprule
\multirow{3}{*}{\textbf{Method}}
& \shortstack[c]{\textbf{Text} \textbf{Align.}}
& \multicolumn{2}{@{} c}{\shortstack[c]{\textbf{Image} \textbf{Align.}}} & \multicolumn{2}{@{} c}{\shortstack[c]{\textbf{Camera-pose} \textbf{Accuracy}}} \\ \cmidrule(lr){2-2} \cmidrule(lr){3-4} \cmidrule(lr){5-6}
 & {\shortstack[c]{CLIP-\\score$\uparrow$}} & {\shortstack[c]{fore-\\ground$\uparrow$}} & {\shortstack[c]{back-\\ground$\downarrow$}} & {\shortstack[c]{ Angular\\ error $\downarrow$}} & {\shortstack[c]{Camera center \\ error $\downarrow$}} \\
\midrule
\textbf{Ours}  &   0.248	&  \textbf{0.471} & 0.348  & 14.19 &  0.080 \\
\textbf{w/o \refeq{featurenerf_update}} & \textbf{0.250} &	\textcolor{gray}{0.460} & \textbf{0.340} & \textcolor{gray}{16.08} &  \textcolor{gray}{0.096} \\
\textbf{w/o $\mathcal{L}_{bg} + \mathcal{L}_{s}$}   & \textcolor{gray}{0.239} & \textbf{0.471} &  \textcolor{gray}{0.371} & \textbf{11.83}	&  \textbf{0.068}  \\
\bottomrule
\end{tabular}
}
\vspace{-8pt}
\caption{\textbf{Ablation experiments.} Not enriching volumetric features with text cross-attention (\refeq{featurenerf_update}) has an adverse effect on image alignment. Not having mask-based losses (\refeq{loss_background}) leads to overfitting on training images and decreases the text alignment. The worst performing metrics are \textcolor{gray}{grayed}. Our final method achieves a balance between the input conditions of the target concept, text prompt, and camera pose. 
}
\label{tbl:ablation_1}
\vspace{-10pt}
\end{table}

\subsection{Ablation}
In this section, we perform ablation experiments regarding different components of our method and show its contribution. All ablation studies are done on CO3D-v2 instances with validation camera poses.

\myparagraph{Background losses.} When removing the silhouette and background loss, as explained in \refeq{loss_background} from training, we observe a decrease in text alignment and overfitting on training images as shown in Table~\ref{tbl:ablation_1}. \reffig{mask_loss} in \refapp{results1} shows qualitatively that the model generates images with backgrounds more similar to the training views. This is also reflected by the higher similarity between generated images and background regions of the training images ($3^{\text{rd}}$ column \reftbl{ablation_1}) compared to our final method.

\myparagraph{Text cross-attention in FeatureNeRF.} We also enrich the 3D learned features with text cross-attention as shown in \refeq{featurenerf_update}. We perform the ablation experiment of removing this component from the module. \reftbl{ablation_1} shows that this leads to a drop in image alignment with the target concept. Thus, cross-attention with text in the volumetric feature space helps the module learn the target concept better. 

\begin{figure}[!t]
    \includegraphics[width=\linewidth]{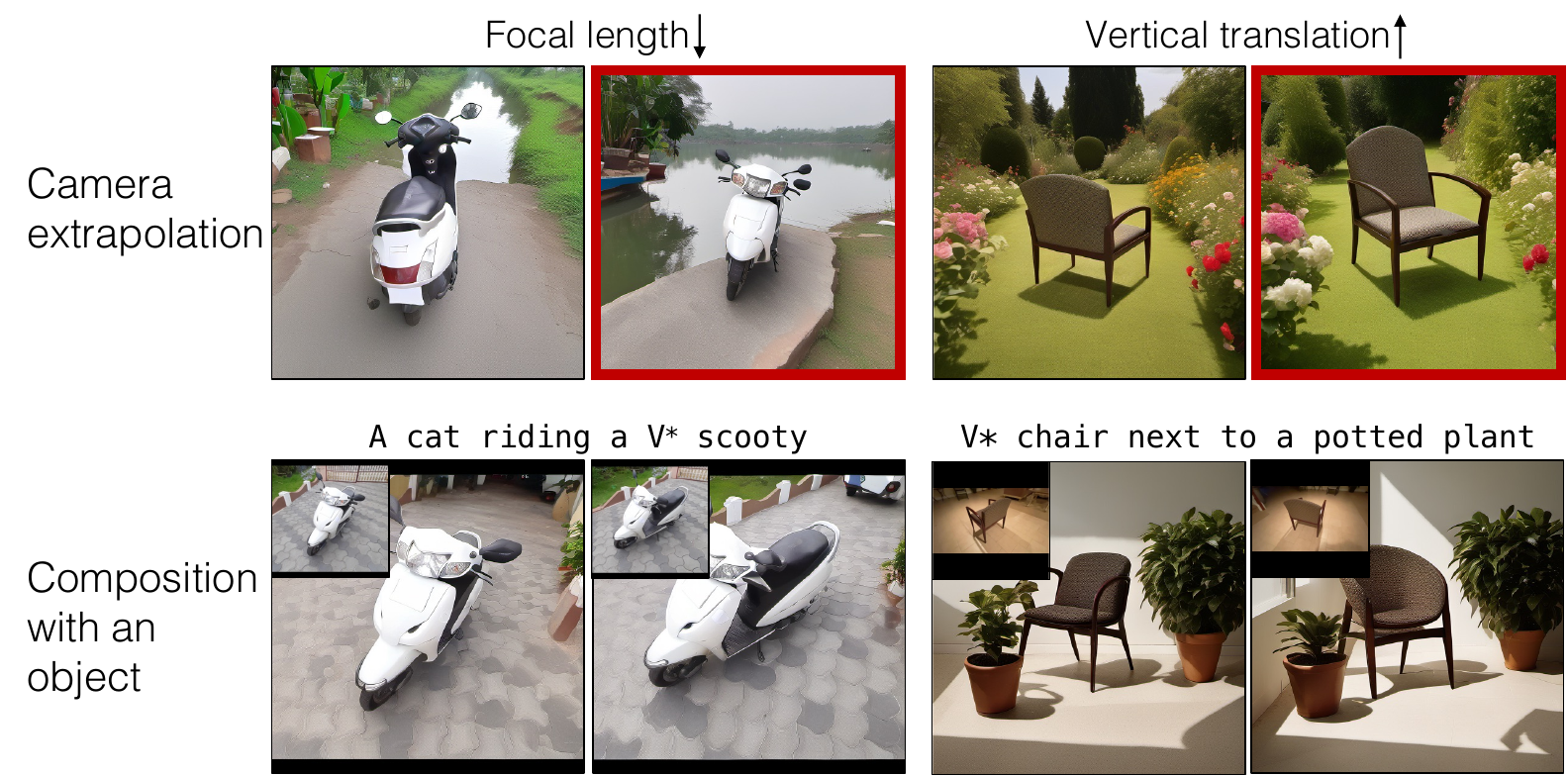}
    \vspace{-15pt}
    \caption{\textbf{Limitations}. Our method can occasionally fail when the target object viewpoint deviates far from the training images, e.g., reducing the focal length too much (top left) or rendering the object off-center (top right), as the pre-trained model is often biased towards generating the object in the center. Also, it can fail to follow the input text prompt or the exact object viewpoint when multiple objects are composed in a scene (bottom row).}
     \vspace{-10pt}
\label{fig:limitation}
\end{figure}

We show more results and ablation experiments in the Appendix, including performance with predicted camera viewpoints.

\section{Discussion and Limitations}
We introduce a new task of customizing text-to-image models with object viewpoint control. Our method learns view-dependent object features in the intermediate feature space of the diffusion model and conditions the generation on them.
This enables synthesizing the object with varying object viewpoints while controlling other aspects through text prompts.

\paragraph{Limitations.} Though our method outperforms existing image editing and model customization approaches, it still has several limitations.
As we show in \reffig{limitation}, our method occasionally struggles at generalizing to extreme viewpoints that were not seen during training and resorts to either changing the object identity or generating the object in a seen viewpoint. We expect this to improve by adding more viewpoint variations during training. Our method also sometimes struggles to follow the input viewpoint condition when the text prompt adds multiple objects to the scene.
We hypothesize that in such challenging scenarios, the model is biased towards generating object-centric front views, as seen in its original training data. Also, we fine-tune the model for each custom object, which takes computation time ($\sim$ 40 minutes). Exploring pose-conditioning in a zero-shot, feed-forward manner~\cite{chen2023subject,gal2023encoder} may help reduce the time and computation. Finally, we focus on enabling viewpoint control for rigid objects. Future work includes extending this conditioning to handle dynamic objects that change the pose in between reference views. One potential way to address this is using a representation based on dynamic and non-rigid NeRF methods~\cite{kplanes_2023, pumarola2020d, song2023totalrecon}.

\myparagraph{Acknowledgment.}
We are thankful to Kangle Deng, Sheng-Yu Wang, and Gaurav Parmar for their helpful comments and discussion and to Sean Liu, Ruihan Gao, Yufei Ye, and Bharath Raj for proofreading the draft. This work was partly done by Nupur Kumari during the Adobe internship. The work is partly supported by Adobe Research, the Packard Fellowship, the Amazon Faculty Research Award, and NSF IIS-2239076. Grace Su is supported by the NSF Graduate Research Fellowship (Grant No. DGE2140739).

{
    \small
    \bibliographystyle{ieeenat_fullname}
    \bibliography{main}

\begin{thebibliography}{111}
\providecommand{\natexlab}[1]{#1}
\providecommand{\url}[1]{\texttt{#1}}
\expandafter\ifx\csname urlstyle\endcsname\relax
  \providecommand{\doi}[1]{doi: #1}\else
  \providecommand{\doi}{doi: \begingroup \urlstyle{rm}\Url}\fi

\bibitem[Adobe(2023)]{generative_fill}
Adobe.
\newblock Generative fill.
\newblock \url{https://www.adobe.com/products/photoshop/generative-fill.html}, 2023.

\bibitem[Alaluf et~al.(2023)Alaluf, Richardson, Metzer, and Cohen-Or]{alaluf2023neural}
Yuval Alaluf, Elad Richardson, Gal Metzer, and Daniel Cohen-Or.
\newblock A neural space-time representation for text-to-image personalization.
\newblock \emph{ACM Transactions on Graphics (TOG)}, 2023.

\bibitem[Arar et~al.(2023)Arar, Gal, Atzmon, Chechik, Cohen-Or, Shamir, and H.~Bermano]{arar2023domain}
Moab Arar, Rinon Gal, Yuval Atzmon, Gal Chechik, Daniel Cohen-Or, Ariel Shamir, and Amit H.~Bermano.
\newblock Domain-agnostic tuning-encoder for fast personalization of text-to-image models.
\newblock In \emph{SIGGRAPH Asia 2023 Conference Papers}, 2023.

\bibitem[Bar-Tal et~al.(2023)Bar-Tal, Yariv, Lipman, and Dekel]{bar2023multidiffusion}
Omer Bar-Tal, Lior Yariv, Yaron Lipman, and Tali Dekel.
\newblock Multidiffusion: Fusing diffusion paths for controlled image generation.
\newblock In \emph{International Conference on Machine Learning (ICML)}, 2023.

\bibitem[Barron et~al.(2021)Barron, Mildenhall, Tancik, Hedman, Martin-Brualla, and Srinivasan]{barron2021mip}
Jonathan~T Barron, Ben Mildenhall, Matthew Tancik, Peter Hedman, Ricardo Martin-Brualla, and Pratul~P Srinivasan.
\newblock Mip-nerf: A multiscale representation for anti-aliasing neural radiance fields.
\newblock In \emph{IEEE International Conference on Computer Vision (ICCV)}, 2021.

\bibitem[Barron et~al.(2023)Barron, Mildenhall, Verbin, Srinivasan, and Hedman]{barron2023zip}
Jonathan~T Barron, Ben Mildenhall, Dor Verbin, Pratul~P Srinivasan, and Peter Hedman.
\newblock Zip-nerf: Anti-aliased grid-based neural radiance fields.
\newblock In \emph{IEEE International Conference on Computer Vision (ICCV)}, 2023.

\bibitem[Boss et~al.(2021)Boss, Braun, Jampani, Barron, Liu, and Lensch]{boss2021nerd}
Mark Boss, Raphael Braun, Varun Jampani, Jonathan~T Barron, Ce Liu, and Hendrik Lensch.
\newblock Nerd: Neural reflectance decomposition from image collections.
\newblock In \emph{IEEE International Conference on Computer Vision (ICCV)}, 2021.

\bibitem[Boss et~al.(2022)Boss, Engelhardt, Kar, Li, Sun, Barron, Lensch, and Jampani]{boss2022samurai}
Mark Boss, Andreas Engelhardt, Abhishek Kar, Yuanzhen Li, Deqing Sun, Jonathan Barron, Hendrik Lensch, and Varun Jampani.
\newblock Samurai: Shape and material from unconstrained real-world arbitrary image collections.
\newblock In \emph{Conference on Neural Information Processing Systems (NeurIPS)}, 2022.

\bibitem[Brack et~al.(2023)Brack, Friedrich, Kornmeier, Tsaban, Schramowski, Kersting, and Passos]{brack2023ledits++}
Manuel Brack, Felix Friedrich, Katharina Kornmeier, Linoy Tsaban, Patrick Schramowski, Kristian Kersting, and Apolin{\'a}rio Passos.
\newblock Ledits++: Limitless image editing using text-to-image models.
\newblock \emph{arXiv preprint arXiv:2311.16711}, 2023.

\bibitem[Brooks et~al.(2023)Brooks, Holynski, and Efros]{brooks2023instructpix2pix}
Tim Brooks, Aleksander Holynski, and Alexei~A Efros.
\newblock Instructpix2pix: Learning to follow image editing instructions.
\newblock In \emph{IEEE Conference on Computer Vision and Pattern Recognition (CVPR)}, 2023.

\bibitem[Burgess et~al.(2024)Burgess, Wang, and Yeung-Levy]{burgess2024viewpointtextualinversiondiscovering}
James Burgess, Kuan-Chieh Wang, and Serena Yeung-Levy.
\newblock Viewpoint textual inversion: Discovering scene representations and 3d view control in 2d diffusion models.
\newblock \emph{European Conference on Computer Vision (ECCV)}, 2024.

\bibitem[Cao et~al.(2023)Cao, Wang, Qi, Shan, Qie, and Zheng]{cao2023masactrl}
Mingdeng Cao, Xintao Wang, Zhongang Qi, Ying Shan, Xiaohu Qie, and Yinqiang Zheng.
\newblock Masactrl: Tuning-free mutual self-attention control for consistent image synthesis and editing.
\newblock In \emph{IEEE International Conference on Computer Vision (ICCV)}, 2023.

\bibitem[Chan et~al.(2023)Chan, Nagano, Chan, Bergman, Park, Levy, Aittala, De~Mello, Karras, and Wetzstein]{chan2023genvs}
Eric~R Chan, Koki Nagano, Matthew~A Chan, Alexander~W Bergman, Jeong~Joon Park, Axel Levy, Miika Aittala, Shalini De~Mello, Tero Karras, and Gordon Wetzstein.
\newblock Genvs: Generative novel view synthesis with 3d-aware diffusion models.
\newblock In \emph{IEEE International Conference on Computer Vision (ICCV)}, 2023.

\bibitem[ChatGPT(2022)]{chatgpt}
ChatGPT.
\newblock Chatgpt.
\newblock \url{https://chat.openai.com/chat}, 2022.

\bibitem[Chefer et~al.(2023)Chefer, Alaluf, Vinker, Wolf, and Cohen-Or]{chefer2023attend}
Hila Chefer, Yuval Alaluf, Yael Vinker, Lior Wolf, and Daniel Cohen-Or.
\newblock Attend-and-excite: Attention-based semantic guidance for text-to-image diffusion models.
\newblock \emph{ACM Transactions on Graphics (TOG)}, 2023.

\bibitem[Chen et~al.(2022)Chen, Xu, Geiger, Yu, and Su]{chen2022tensorf}
Anpei Chen, Zexiang Xu, Andreas Geiger, Jingyi Yu, and Hao Su.
\newblock Tensorf: Tensorial radiance fields.
\newblock In \emph{European Conference on Computer Vision (ECCV)}, 2022.

\bibitem[Chen et~al.(2013)Chen, Zhu, Shamir, Hu, and Cohen-Or]{chen20133}
Tao Chen, Zhe Zhu, Ariel Shamir, Shi-Min Hu, and Daniel Cohen-Or.
\newblock 3-sweep: Extracting editable objects from a single photo.
\newblock \emph{ACM Transactions on graphics (TOG)}, 2013.

\bibitem[Chen et~al.(2023)Chen, Hu, Li, Rui, Jia, Chang, and Cohen]{chen2023subject}
Wenhu Chen, Hexiang Hu, Yandong Li, Nataniel Rui, Xuhui Jia, Ming-Wei Chang, and William~W Cohen.
\newblock Subject-driven text-to-image generation via apprenticeship learning.
\newblock In \emph{Conference on Neural Information Processing Systems (NeurIPS)}, 2023.

\bibitem[Cheng et~al.(2024)Cheng, Gadelha, Groueix, Fisher, Mech, Markham, and Trigoni]{cheng2024learning}
Ta-Ying Cheng, Matheus Gadelha, Thibault Groueix, Matthew Fisher, Radomir Mech, Andrew Markham, and Niki Trigoni.
\newblock Learning continuous 3d words for text-to-image generation.
\newblock In \emph{IEEE Conference on Computer Vision and Pattern Recognition (CVPR)}, 2024.

\bibitem[Deng et~al.(2022)Deng, Liu, Zhu, and Ramanan]{deng2022depth}
Kangle Deng, Andrew Liu, Jun-Yan Zhu, and Deva Ramanan.
\newblock Depth-supervised nerf: Fewer views and faster training for free.
\newblock In \emph{IEEE Conference on Computer Vision and Pattern Recognition (CVPR)}, 2022.

\bibitem[Dhariwal and Nichol(2021)]{dhariwal2021diffusion}
Prafulla Dhariwal and Alexander Nichol.
\newblock Diffusion models beat gans on image synthesis.
\newblock In \emph{Conference on Neural Information Processing Systems (NeurIPS)}, 2021.

\bibitem[Dong and Wang(2023)]{dong2024vica}
Jiahua Dong and Yu-Xiong Wang.
\newblock Vica-nerf: View-consistency-aware 3d editing of neural radiance fields.
\newblock In \emph{Conference on Neural Information Processing Systems (NeurIPS)}, 2023.

\bibitem[Fridovich-Keil et~al.(2023)Fridovich-Keil, Meanti, Warburg, Recht, and Kanazawa]{kplanes_2023}
Sara Fridovich-Keil, Giacomo Meanti, Frederik~Rahbæk Warburg, Benjamin Recht, and Angjoo Kanazawa.
\newblock K-planes: Explicit radiance fields in space, time, and appearance.
\newblock In \emph{IEEE Conference on Computer Vision and Pattern Recognition (CVPR)}, 2023.

\bibitem[Gafni et~al.(2022)Gafni, Polyak, Ashual, Sheynin, Parikh, and Taigman]{gafni2022make}
Oran Gafni, Adam Polyak, Oron Ashual, Shelly Sheynin, Devi Parikh, and Yaniv Taigman.
\newblock Make-a-scene: Scene-based text-to-image generation with human priors.
\newblock In \emph{European Conference on Computer Vision (ECCV)}, 2022.

\bibitem[Gal et~al.(2023{\natexlab{a}})Gal, Alaluf, Atzmon, Patashnik, Bermano, Chechik, and Cohen-Or]{gal2022image}
Rinon Gal, Yuval Alaluf, Yuval Atzmon, Or Patashnik, Amit~H Bermano, Gal Chechik, and Daniel Cohen-Or.
\newblock An image is worth one word: Personalizing text-to-image generation using textual inversion.
\newblock In \emph{International Conference on Learning Representations (ICLR)}, 2023{\natexlab{a}}.

\bibitem[Gal et~al.(2023{\natexlab{b}})Gal, Arar, Atzmon, Bermano, Chechik, and Cohen-Or]{gal2023encoder}
Rinon Gal, Moab Arar, Yuval Atzmon, Amit~H Bermano, Gal Chechik, and Daniel Cohen-Or.
\newblock Encoder-based domain tuning for fast personalization of text-to-image models.
\newblock \emph{ACM Transactions on Graphics (TOG)}, 2023{\natexlab{b}}.

\bibitem[Ge et~al.(2023)Ge, Park, Zhu, and Huang]{ge2023expressive}
Songwei Ge, Taesung Park, Jun-Yan Zhu, and Jia-Bin Huang.
\newblock Expressive text-to-image generation with rich text.
\newblock In \emph{IEEE International Conference on Computer Vision (ICCV)}, 2023.

\bibitem[Han et~al.(2023)Han, Li, Zhang, Milanfar, Metaxas, and Yang]{han2023svdiff}
Ligong Han, Yinxiao Li, Han Zhang, Peyman Milanfar, Dimitris Metaxas, and Feng Yang.
\newblock Svdiff: Compact parameter space for diffusion fine-tuning.
\newblock In \emph{IEEE International Conference on Computer Vision (ICCV)}, 2023.

\bibitem[Haque et~al.(2023)Haque, Tancik, Efros, Holynski, and Kanazawa]{haque2023instruct}
Ayaan Haque, Matthew Tancik, Alexei~A Efros, Aleksander Holynski, and Angjoo Kanazawa.
\newblock Instruct-nerf2nerf: Editing 3d scenes with instructions.
\newblock In \emph{IEEE International Conference on Computer Vision (ICCV)}, 2023.

\bibitem[He et~al.(2016)He, Zhang, Ren, and Sun]{resnet}
Kaiming He, Xiangyu Zhang, Shaoqing Ren, and Jian Sun.
\newblock Deep residual learning for image recognition.
\newblock In \emph{IEEE Conference on Computer Vision and Pattern Recognition (CVPR)}, 2016.

\bibitem[Hertz et~al.(2023)Hertz, Mokady, Tenenbaum, Aberman, Pritch, and Cohen-Or]{hertz2022prompt}
Amir Hertz, Ron Mokady, Jay Tenenbaum, Kfir Aberman, Yael Pritch, and Daniel Cohen-Or.
\newblock Prompt-to-prompt image editing with cross attention control.
\newblock In \emph{International Conference on Learning Representations (ICLR)}, 2023.

\bibitem[Ho et~al.(2020)Ho, Jain, and Abbeel]{ho2020denoising}
Jonathan Ho, Ajay Jain, and Pieter Abbeel.
\newblock Denoising diffusion probabilistic models.
\newblock In \emph{Conference on Neural Information Processing Systems (NeurIPS)}, 2020.

\bibitem[H{\"o}llein et~al.(2024)H{\"o}llein, Bovzivc, M{\"u}ller, Novotny, Tseng, Richardt, Zollh{\"o}fer, and Nie{\ss}ner]{hollein2024viewdiff}
Lukas H{\"o}llein, Aljavz Bovzivc, Norman M{\"u}ller, David Novotny, Hung-Yu Tseng, Christian Richardt, Michael Zollh{\"o}fer, and Matthias Nie{\ss}ner.
\newblock Viewdiff: 3d-consistent image generation with text-to-image models.
\newblock In \emph{IEEE Conference on Computer Vision and Pattern Recognition (CVPR)}, 2024.

\bibitem[Hu et~al.(2022)Hu, Shen, Wallis, Allen-Zhu, Li, Wang, Wang, and Chen]{hu2021lora}
Edward~J Hu, Yelong Shen, Phillip Wallis, Zeyuan Allen-Zhu, Yuanzhi Li, Shean Wang, Lu Wang, and Weizhu Chen.
\newblock Lora: Low-rank adaptation of large language models.
\newblock In \emph{International Conference on Learning Representations (ICLR)}, 2022.

\bibitem[Jampani et~al.(2023)Jampani, Maninis, Engelhardt, Karpur, Truong, Sargent, Popov, Araujo, Martin~Brualla, Patel, et~al.]{jampani2024navi}
Varun Jampani, Kevis-Kokitsi Maninis, Andreas Engelhardt, Arjun Karpur, Karen Truong, Kyle Sargent, Stefan Popov, Andr{\'e} Araujo, Ricardo Martin~Brualla, Kaushal Patel, et~al.
\newblock Navi: Category-agnostic image collections with high-quality 3d shape and pose annotations.
\newblock In \emph{Conference on Neural Information Processing Systems (NeurIPS)}, 2023.

\bibitem[Kang et~al.(2023)Kang, Zhu, Zhang, Park, Shechtman, Paris, and Park]{kang2023scaling}
Minguk Kang, Jun-Yan Zhu, Richard Zhang, Jaesik Park, Eli Shechtman, Sylvain Paris, and Taesung Park.
\newblock Scaling up gans for text-to-image synthesis.
\newblock In \emph{IEEE Conference on Computer Vision and Pattern Recognition (CVPR)}, 2023.

\bibitem[Karras et~al.(2022)Karras, Aittala, Aila, and Laine]{karras2022elucidating}
Tero Karras, Miika Aittala, Timo Aila, and Samuli Laine.
\newblock Elucidating the design space of diffusion-based generative models.
\newblock In \emph{Conference on Neural Information Processing Systems (NeurIPS)}, 2022.

\bibitem[Karras et~al.(2023)Karras, Aittala, Lehtinen, Hellsten, Aila, and Laine]{karras2023analyzing}
Tero Karras, Miika Aittala, Jaakko Lehtinen, Janne Hellsten, Timo Aila, and Samuli Laine.
\newblock Analyzing and improving the training dynamics of diffusion models.
\newblock \emph{arXiv preprint arXiv:2312.02696}, 2023.

\bibitem[Karsch et~al.(2011)Karsch, Hedau, Forsyth, and Hoiem]{karsch2011rendering}
Kevin Karsch, Varsha Hedau, David Forsyth, and Derek Hoiem.
\newblock Rendering synthetic objects into legacy photographs.
\newblock \emph{ACM Transactions on graphics (TOG)}, 2011.

\bibitem[Kawar et~al.(2023)Kawar, Zada, Lang, Tov, Chang, Dekel, Mosseri, and Irani]{kawar2023imagic}
Bahjat Kawar, Shiran Zada, Oran Lang, Omer Tov, Huiwen Chang, Tali Dekel, Inbar Mosseri, and Michal Irani.
\newblock Imagic: Text-based real image editing with diffusion models.
\newblock In \emph{IEEE Conference on Computer Vision and Pattern Recognition (CVPR)}, 2023.

\bibitem[Kerbl et~al.(2023)Kerbl, Kopanas, Leimk{\"u}hler, and Drettakis]{kerbl20233d}
Bernhard Kerbl, Georgios Kopanas, Thomas Leimk{\"u}hler, and George Drettakis.
\newblock 3d gaussian splatting for real-time radiance field rendering.
\newblock \emph{ACM Transactions on Graphics}, 42\penalty0 (4), 2023.

\bibitem[Kerr et~al.(2023)Kerr, Kim, Goldberg, Kanazawa, and Tancik]{kerr2023lerf}
Justin Kerr, Chung~Min Kim, Ken Goldberg, Angjoo Kanazawa, and Matthew Tancik.
\newblock Lerf: Language embedded radiance fields.
\newblock In \emph{IEEE International Conference on Computer Vision (ICCV)}, 2023.

\bibitem[Kholgade et~al.(2014)Kholgade, Simon, Efros, and Sheikh]{kholgade20143d}
Natasha Kholgade, Tomas Simon, Alexei Efros, and Yaser Sheikh.
\newblock 3d object manipulation in a single photograph using stock 3d models.
\newblock \emph{ACM Transactions on graphics (TOG)}, 2014.

\bibitem[Kim et~al.(2023)Kim, Lee, Kim, Ha, and Zhu]{densediffusion}
Yunji Kim, Jiyoung Lee, Jin-Hwa Kim, Jung-Woo Ha, and Jun-Yan Zhu.
\newblock Dense text-to-image generation with attention modulation.
\newblock In \emph{IEEE International Conference on Computer Vision (ICCV)}, 2023.

\bibitem[Kingma and Welling(2014)]{kingma2013auto}
Diederik~P Kingma and Max Welling.
\newblock Auto-encoding variational bayes.
\newblock In \emph{International Conference on Learning Representations (ICLR)}, 2014.

\bibitem[Kirillov et~al.(2023)Kirillov, Mintun, Ravi, Mao, Rolland, Gustafson, Xiao, Whitehead, Berg, Lo, et~al.]{kirillov2023segment}
Alexander Kirillov, Eric Mintun, Nikhila Ravi, Hanzi Mao, Chloe Rolland, Laura Gustafson, Tete Xiao, Spencer Whitehead, Alexander~C Berg, Wan-Yen Lo, et~al.
\newblock Segment anything.
\newblock In \emph{IEEE International Conference on Computer Vision (ICCV)}, 2023.

\bibitem[Kumari et~al.(2023)Kumari, Zhang, Zhang, Shechtman, and Zhu]{kumari2023multi}
Nupur Kumari, Bingliang Zhang, Richard Zhang, Eli Shechtman, and Jun-Yan Zhu.
\newblock Multi-concept customization of text-to-image diffusion.
\newblock In \emph{IEEE Conference on Computer Vision and Pattern Recognition (CVPR)}, 2023.

\bibitem[Li et~al.(2023)Li, Li, and Hoi]{li2023blip}
Dongxu Li, Junnan Li, and Steven~CH Hoi.
\newblock Blip-diffusion: Pre-trained subject representation for controllable text-to-image generation and editing.
\newblock In \emph{Conference on Neural Information Processing Systems (NeurIPS)}, 2023.

\bibitem[Liu et~al.(2022)Liu, Ren, Lin, and Zhao]{liu2022pseudo}
Luping Liu, Yi Ren, Zhijie Lin, and Zhou Zhao.
\newblock Pseudo numerical methods for diffusion models on manifolds.
\newblock In \emph{International Conference on Learning Representations (ICLR)}, 2022.

\bibitem[Liu et~al.(2023)Liu, Wu, Van~Hoorick, Tokmakov, Zakharov, and Vondrick]{liu2023zero}
Ruoshi Liu, Rundi Wu, Basile Van~Hoorick, Pavel Tokmakov, Sergey Zakharov, and Carl Vondrick.
\newblock Zero-1-to-3: Zero-shot one image to 3d object.
\newblock In \emph{IEEE International Conference on Computer Vision (ICCV)}, 2023.

\bibitem[Liu et~al.(2024)Liu, Lin, Zeng, Long, Liu, Komura, and Wang]{liu2023syncdreamer}
Yuan Liu, Cheng Lin, Zijiao Zeng, Xiaoxiao Long, Lingjie Liu, Taku Komura, and Wenping Wang.
\newblock Syncdreamer: Generating multiview-consistent images from a single-view image.
\newblock In \emph{International Conference on Learning Representations (ICLR)}, 2024.

\bibitem[Lu et~al.(2022)Lu, Zhou, Bao, Chen, Li, and Zhu]{lu2022dpm}
Cheng Lu, Yuhao Zhou, Fan Bao, Jianfei Chen, Chongxuan Li, and Jun Zhu.
\newblock Dpm-solver: A fast ode solver for diffusion probabilistic model sampling in around 10 steps.
\newblock In \emph{Conference on Neural Information Processing Systems (NeurIPS)}, 2022.

\bibitem[Meng et~al.(2022)Meng, He, Song, Song, Wu, Zhu, and Ermon]{meng2021sdedit}
Chenlin Meng, Yutong He, Yang Song, Jiaming Song, Jiajun Wu, Jun-Yan Zhu, and Stefano Ermon.
\newblock Sdedit: Guided image synthesis and editing with stochastic differential equations.
\newblock In \emph{International Conference on Learning Representations (ICLR)}, 2022.

\bibitem[Metzer et~al.(2023)Metzer, Richardson, Patashnik, Giryes, and Cohen-Or]{metzer2023latent}
Gal Metzer, Elad Richardson, Or Patashnik, Raja Giryes, and Daniel Cohen-Or.
\newblock Latent-nerf for shape-guided generation of 3d shapes and textures.
\newblock In \emph{IEEE Conference on Computer Vision and Pattern Recognition (CVPR)}, 2023.

\bibitem[Michel et~al.(2023)Michel, Bhattad, VanderBilt, Krishna, Kembhavi, and Gupta]{michel2023object}
Oscar Michel, Anand Bhattad, Eli VanderBilt, Ranjay Krishna, Aniruddha Kembhavi, and Tanmay Gupta.
\newblock Object 3dit: Language-guided 3d-aware image editing.
\newblock In \emph{Conference on Neural Information Processing Systems (NeurIPS)}, 2023.

\bibitem[Mildenhall et~al.(2021)Mildenhall, Srinivasan, Tancik, Barron, Ramamoorthi, and Ng]{mildenhall2021nerf}
Ben Mildenhall, Pratul~P Srinivasan, Matthew Tancik, Jonathan~T Barron, Ravi Ramamoorthi, and Ren Ng.
\newblock Nerf: Representing scenes as neural radiance fields for view synthesis.
\newblock \emph{Communications of the ACM}, 2021.

\bibitem[Mokady et~al.(2023)Mokady, Hertz, Aberman, Pritch, and Cohen-Or]{mokady2023null}
Ron Mokady, Amir Hertz, Kfir Aberman, Yael Pritch, and Daniel Cohen-Or.
\newblock Null-text inversion for editing real images using guided diffusion models.
\newblock In \emph{IEEE Conference on Computer Vision and Pattern Recognition (CVPR)}, 2023.

\bibitem[Mou et~al.(2024)Mou, Wang, Xie, Wu, Zhang, Qi, and Shan]{mou2024t2i}
Chong Mou, Xintao Wang, Liangbin Xie, Yanze Wu, Jian Zhang, Zhongang Qi, and Ying Shan.
\newblock T2i-adapter: Learning adapters to dig out more controllable ability for text-to-image diffusion models.
\newblock In \emph{Conference on Artificial Intelligence (AAAI)}, 2024.

\bibitem[M{\"u}ller et~al.(2022)M{\"u}ller, Evans, Schied, and Keller]{muller2022instant}
Thomas M{\"u}ller, Alex Evans, Christoph Schied, and Alexander Keller.
\newblock Instant neural graphics primitives with a multiresolution hash encoding.
\newblock \emph{ACM Transactions on Graphics (ToG)}, 2022.

\bibitem[Niemeyer et~al.(2022)Niemeyer, Barron, Mildenhall, Sajjadi, Geiger, and Radwan]{Niemeyer2021Regnerf}
Michael Niemeyer, Jonathan~T. Barron, Ben Mildenhall, Mehdi S.~M. Sajjadi, Andreas Geiger, and Noha Radwan.
\newblock Regnerf: Regularizing neural radiance fields for view synthesis from sparse inputs.
\newblock In \emph{IEEE Conference on Computer Vision and Pattern Recognition (CVPR)}, 2022.

\bibitem[Oquab et~al.(2023)Oquab, Darcet, Moutakanni, Vo, Szafraniec, Khalidov, Fernandez, Haziza, Massa, El-Nouby, et~al.]{oquab2023dinov2}
Maxime Oquab, Timoth{\'e}e Darcet, Th{\'e}o Moutakanni, Huy Vo, Marc Szafraniec, Vasil Khalidov, Pierre Fernandez, Daniel Haziza, Francisco Massa, Alaaeldin El-Nouby, et~al.
\newblock Dinov2: Learning robust visual features without supervision.
\newblock In \emph{TMLR}, 2023.

\bibitem[Parmar et~al.(2023)Parmar, Kumar~Singh, Zhang, Li, Lu, and Zhu]{parmar2023zero}
Gaurav Parmar, Krishna Kumar~Singh, Richard Zhang, Yijun Li, Jingwan Lu, and Jun-Yan Zhu.
\newblock Zero-shot image-to-image translation.
\newblock In \emph{ACM SIGGRAPH 2023 Conference Proceedings}, pages 1--11, 2023.

\bibitem[Patashnik et~al.(2023)Patashnik, Garibi, Azuri, Averbuch-Elor, and Cohen-Or]{patashnik2023localizing}
Or Patashnik, Daniel Garibi, Idan Azuri, Hadar Averbuch-Elor, and Daniel Cohen-Or.
\newblock Localizing object-level shape variations with text-to-image diffusion models.
\newblock In \emph{IEEE International Conference on Computer Vision (ICCV)}, 2023.

\bibitem[Peebles and Xie(2023)]{peebles2023scalable}
William Peebles and Saining Xie.
\newblock Scalable diffusion models with transformers.
\newblock In \emph{IEEE International Conference on Computer Vision (ICCV)}, 2023.

\bibitem[Podell et~al.(2023)Podell, English, Lacey, Blattmann, Dockhorn, M{\"u}ller, Penna, and Rombach]{podell2023sdxl}
Dustin Podell, Zion English, Kyle Lacey, Andreas Blattmann, Tim Dockhorn, Jonas M{\"u}ller, Joe Penna, and Robin Rombach.
\newblock Sdxl: Improving latent diffusion models for high-resolution image synthesis.
\newblock \emph{arXiv preprint arXiv:2307.01952}, 2023.

\bibitem[Pumarola et~al.(2021)Pumarola, Corona, Pons-Moll, and Moreno-Noguer]{pumarola2020d}
Albert Pumarola, Enric Corona, Gerard Pons-Moll, and Francesc Moreno-Noguer.
\newblock {D-NeRF: Neural Radiance Fields for Dynamic Scenes}.
\newblock In \emph{IEEE Conference on Computer Vision and Pattern Recognition (CVPR)}, 2021.

\bibitem[Radford et~al.(2021)Radford, Kim, Hallacy, Ramesh, Goh, Agarwal, Sastry, Askell, Mishkin, Clark, et~al.]{radford2021learning}
Alec Radford, Jong~Wook Kim, Chris Hallacy, Aditya Ramesh, Gabriel Goh, Sandhini Agarwal, Girish Sastry, Amanda Askell, Pamela Mishkin, Jack Clark, et~al.
\newblock Learning transferable visual models from natural language supervision.
\newblock In \emph{International Conference on Machine Learning (ICML)}, 2021.

\bibitem[Raj et~al.(2023)Raj, Kaza, Poole, Niemeyer, Ruiz, Mildenhall, Zada, Aberman, Rubinstein, Barron, et~al.]{raj2023dreambooth3d}
Amit Raj, Srinivas Kaza, Ben Poole, Michael Niemeyer, Nataniel Ruiz, Ben Mildenhall, Shiran Zada, Kfir Aberman, Michael Rubinstein, Jonathan Barron, et~al.
\newblock Dreambooth3d: Subject-driven text-to-3d generation.
\newblock In \emph{IEEE International Conference on Computer Vision (ICCV)}, 2023.

\bibitem[Ramesh et~al.(2022)Ramesh, Dhariwal, Nichol, Chu, and Chen]{ramesh2022hierarchical}
Aditya Ramesh, Prafulla Dhariwal, Alex Nichol, Casey Chu, and Mark Chen.
\newblock Hierarchical text-conditional image generation with clip latents.
\newblock \emph{arXiv preprint arXiv:2204.06125}, 2022.

\bibitem[Ravi et~al.(2020)Ravi, Reizenstein, Novotny, Gordon, Lo, Johnson, and Gkioxari]{ravi2020accelerating}
Nikhila Ravi, Jeremy Reizenstein, David Novotny, Taylor Gordon, Wan-Yen Lo, Justin Johnson, and Georgia Gkioxari.
\newblock Accelerating 3d deep learning with pytorch3d.
\newblock \emph{arXiv preprint arXiv:2007.08501}, 2020.

\bibitem[Reizenstein et~al.(2021)Reizenstein, Shapovalov, Henzler, Sbordone, Labatut, Taigman, and Novotny]{reizenstein21co3d}
Jeremy Reizenstein, Roman Shapovalov, Philipp Henzler, Luca Sbordone, Devi Labatut, Patrikh, Yanivck Taigman, and David Novotny.
\newblock Common objects in 3d: Large-scale learning and evaluation of real-life 3d category reconstruction.
\newblock In \emph{IEEE International Conference on Computer Vision (ICCV)}, 2021.

\bibitem[Rombach et~al.(2022)Rombach, Blattmann, Lorenz, Esser, and Ommer]{rombach2022high}
Robin Rombach, Andreas Blattmann, Dominik Lorenz, Patrick Esser, and Bj{\"o}rn Ommer.
\newblock High-resolution image synthesis with latent diffusion models.
\newblock In \emph{IEEE Conference on Computer Vision and Pattern Recognition (CVPR)}, 2022.

\bibitem[Ronneberger et~al.(2015)Ronneberger, Fischer, and Brox]{ronneberger2015u}
Olaf Ronneberger, Philipp Fischer, and Thomas Brox.
\newblock U-net: Convolutional networks for biomedical image segmentation.
\newblock In \emph{International Conference on Medical image computing and computer-assisted intervention}, 2015.

\bibitem[Ruiz et~al.(2023{\natexlab{a}})Ruiz, Li, Jampani, Pritch, Rubinstein, and Aberman]{ruiz2022dreambooth}
Nataniel Ruiz, Yuanzhen Li, Varun Jampani, Yael Pritch, Michael Rubinstein, and Kfir Aberman.
\newblock Dreambooth: Fine tuning text-to-image diffusion models for subject-driven generation.
\newblock In \emph{IEEE Conference on Computer Vision and Pattern Recognition (CVPR)}, 2023{\natexlab{a}}.

\bibitem[Ruiz et~al.(2023{\natexlab{b}})Ruiz, Li, Jampani, Wei, Hou, Pritch, Wadhwa, Rubinstein, and Aberman]{ruiz2023hyperdreambooth}
Nataniel Ruiz, Yuanzhen Li, Varun Jampani, Wei Wei, Tingbo Hou, Yael Pritch, Neal Wadhwa, Michael Rubinstein, and Kfir Aberman.
\newblock Hyperdreambooth: Hypernetworks for fast personalization of text-to-image models.
\newblock \emph{arXiv preprint arXiv:2307.06949}, 2023{\natexlab{b}}.

\bibitem[Ryu(2023)]{loraimplementation}
Simo Ryu.
\newblock Lora-stable diffusion.
\newblock \url{https://github.com/cloneofsimo/lora}, 2023.

\bibitem[Saharia et~al.(2022)Saharia, Chan, Saxena, Li, Whang, Denton, Ghasemipour, Ayan, Mahdavi, Lopes, et~al.]{saharia2022photorealistic}
Chitwan Saharia, William Chan, Saurabh Saxena, Lala Li, Jay Whang, Emily Denton, Seyed Kamyar~Seyed Ghasemipour, Burcu~Karagol Ayan, S~Sara Mahdavi, Rapha~Gontijo Lopes, et~al.
\newblock Photorealistic text-to-image diffusion models with deep language understanding.
\newblock In \emph{Conference on Neural Information Processing Systems (NeurIPS)}, 2022.

\bibitem[Sargent et~al.(2023)Sargent, Li, Shah, Herrmann, Yu, Zhang, Chan, Lagun, Fei-Fei, Sun, et~al.]{sargent2023zeronvs}
Kyle Sargent, Zizhang Li, Tanmay Shah, Charles Herrmann, Hong-Xing Yu, Yunzhi Zhang, Eric~Ryan Chan, Dmitry Lagun, Li Fei-Fei, Deqing Sun, et~al.
\newblock Zeronvs: Zero-shot 360-degree view synthesis from a single real image.
\newblock \emph{arXiv preprint arXiv:2310.17994}, 2023.

\bibitem[Sauer et~al.(2023)Sauer, Karras, Laine, Geiger, and Aila]{sauer2023stylegan}
Axel Sauer, Tero Karras, Samuli Laine, Andreas Geiger, and Timo Aila.
\newblock Stylegan-t: Unlocking the power of gans for fast large-scale text-to-image synthesis.
\newblock In \emph{International Conference on Machine Learning (ICML)}, 2023.

\bibitem[Schuhmann et~al.(2021)Schuhmann, Vencu, Beaumont, Kaczmarczyk, Mullis, Katta, Coombes, Jitsev, and Komatsuzaki]{schuhmann2021laion}
Christoph Schuhmann, Richard Vencu, Romain Beaumont, Robert Kaczmarczyk, Clayton Mullis, Aarush Katta, Theo Coombes, Jenia Jitsev, and Aran Komatsuzaki.
\newblock Laion-400m: Open dataset of clip-filtered 400 million image-text pairs.
\newblock \emph{arXiv preprint arXiv:2111.02114}, 2021.

\bibitem[Shi et~al.(2023)Shi, Xiong, Lin, and Jung]{shi2023instantbooth}
Jing Shi, Wei Xiong, Zhe Lin, and Hyun~Joon Jung.
\newblock Instantbooth: Personalized text-to-image generation without test-time finetuning.
\newblock \emph{arXiv preprint arXiv:2304.03411}, 2023.

\bibitem[Shi et~al.(2024)Shi, Wang, Ye, Long, Li, and Yang]{shi2023mvdream}
Yichun Shi, Peng Wang, Jianglong Ye, Mai Long, Kejie Li, and Xiao Yang.
\newblock Mvdream: Multi-view diffusion for 3d generation.
\newblock In \emph{International Conference on Learning Representations (ICLR)}, 2024.

\bibitem[Sohl-Dickstein et~al.(2015)Sohl-Dickstein, Weiss, Maheswaranathan, and Ganguli]{sohl2015deep}
Jascha Sohl-Dickstein, Eric Weiss, Niru Maheswaranathan, and Surya Ganguli.
\newblock Deep unsupervised learning using nonequilibrium thermodynamics.
\newblock In \emph{International Conference on Machine Learning (ICML)}, 2015.

\bibitem[Song et~al.(2023)Song, Yang, Deng, Zhu, and Ramanan]{song2023totalrecon}
Chonghyuk Song, Gengshan Yang, Kangle Deng, Jun-Yan Zhu, and Deva Ramanan.
\newblock Total-recon: Deformable scene reconstruction for embodied view synthesis.
\newblock In \emph{IEEE International Conference on Computer Vision (ICCV)}, 2023.

\bibitem[Song et~al.(2021)Song, Meng, and Ermon]{song2020denoising}
Jiaming Song, Chenlin Meng, and Stefano Ermon.
\newblock Denoising diffusion implicit models.
\newblock In \emph{International Conference on Learning Representations (ICLR)}, 2021.

\bibitem[Tancik et~al.(2021)Tancik, Mildenhall, Wang, Schmidt, Srinivasan, Barron, and Ng]{tancik2021learned}
Matthew Tancik, Ben Mildenhall, Terrance Wang, Divi Schmidt, Pratul~P Srinivasan, Jonathan~T Barron, and Ren Ng.
\newblock Learned initializations for optimizing coordinate-based neural representations.
\newblock In \emph{IEEE Conference on Computer Vision and Pattern Recognition (CVPR)}, 2021.

\bibitem[Tancik et~al.(2023)Tancik, Weber, Ng, Li, Yi, Wang, Kristoffersen, Austin, Salahi, Ahuja, et~al.]{tancik2023nerfstudio}
Matthew Tancik, Ethan Weber, Evonne Ng, Ruilong Li, Brent Yi, Terrance Wang, Alexander Kristoffersen, Jake Austin, Kamyar Salahi, Abhik Ahuja, et~al.
\newblock Nerfstudio: A modular framework for neural radiance field development.
\newblock In \emph{SIGGRAPH 2023 Conference Papers}, 2023.

\bibitem[Tang et~al.(2024)Tang, Ren, Zhou, Liu, and Zeng]{tang2023dreamgaussian}
Jiaxiang Tang, Jiawei Ren, Hang Zhou, Ziwei Liu, and Gang Zeng.
\newblock Dreamgaussian: Generative gaussian splatting for efficient 3d content creation.
\newblock In \emph{ICLR}, 2024.

\bibitem[Tewel et~al.(2023)Tewel, Gal, Chechik, and Atzmon]{tewel2023key}
Yoad Tewel, Rinon Gal, Gal Chechik, and Yuval Atzmon.
\newblock Key-locked rank one editing for text-to-image personalization.
\newblock In \emph{ACM SIGGRAPH 2023 Conference Proceedings}, 2023.

\bibitem[Valevski et~al.(2023)Valevski, Lumen, Matias, and Leviathan]{valevski2023face0}
Dani Valevski, Danny Lumen, Yossi Matias, and Yaniv Leviathan.
\newblock Face0: Instantaneously conditioning a text-to-image model on a face.
\newblock In \emph{SIGGRAPH Asia 2023 Conference Papers}, 2023.

\bibitem[Vaswani et~al.(2017)Vaswani, Shazeer, Parmar, Uszkoreit, Jones, Gomez, Kaiser, and Polosukhin]{vaswani2017attention}
Ashish Vaswani, Noam Shazeer, Niki Parmar, Jakob Uszkoreit, Llion Jones, Aidan~N Gomez, Lukasz Kaiser, and Illia Polosukhin.
\newblock Attention is all you need.
\newblock In \emph{Conference on Neural Information Processing Systems (NeurIPS)}, 2017.

\bibitem[Voynov et~al.(2023)Voynov, Chu, Cohen-Or, and Aberman]{voynov2023p+}
Andrey Voynov, Qinghao Chu, Daniel Cohen-Or, and Kfir Aberman.
\newblock $ p+ $: Extended textual conditioning in text-to-image generation.
\newblock \emph{arXiv}, 2023.

\bibitem[Wang et~al.(2023)Wang, Efros, Zhu, and Zhang]{wang2023evaluating}
Sheng-Yu Wang, Alexei~A Efros, Jun-Yan Zhu, and Richard Zhang.
\newblock Evaluating data attribution for text-to-image models.
\newblock In \emph{IEEE International Conference on Computer Vision (ICCV)}, 2023.

\bibitem[Wei et~al.(2023)Wei, Zhang, Ji, Bai, Zhang, and Zuo]{wei2023elite}
Yuxiang Wei, Yabo Zhang, Zhilong Ji, Jinfeng Bai, Lei Zhang, and Wangmeng Zuo.
\newblock Elite: Encoding visual concepts into textual embeddings for customized text-to-image generation.
\newblock In \emph{IEEE International Conference on Computer Vision (ICCV)}, 2023.

\bibitem[Wu et~al.(2023)Wu, Mildenhall, Henzler, Park, Gao, Watson, Srinivasan, Verbin, Barron, Poole, et~al.]{wu2023reconfusion}
Rundi Wu, Ben Mildenhall, Philipp Henzler, Keunhong Park, Ruiqi Gao, Daniel Watson, Pratul~P Srinivasan, Dor Verbin, Jonathan~T Barron, Ben Poole, et~al.
\newblock Reconfusion: 3d reconstruction with diffusion priors.
\newblock \emph{arXiv preprint arXiv:2312.02981}, 2023.

\bibitem[Xu et~al.(2023)Xu, Chai, Shi, Peng, Skorokhodov, Siarohin, Yang, Shen, Lee, Zhou, and Tulyakov]{discoscene_2023}
Yinghao Xu, Menglei Chai, Zifan Shi, Sida Peng, Ivan Skorokhodov, Aliaksandr Siarohin, Ceyuan Yang, Yujun Shen, Hsin-Ying Lee, Bolei Zhou, and Sergey Tulyakov.
\newblock Discoscene: Spatially disentangled generative radiance fields for controllable 3d-aware scene synthesis.
\newblock In \emph{IEEE Conference on Computer Vision and Pattern Recognition (CVPR)}, 2023.

\bibitem[Xu et~al.(2024)Xu, Tan, Luan, Bi, Wang, Li, Shi, Sunkavalli, Wetzstein, Xu, and Zhang]{xu2023dmv3d}
Yinghao Xu, Hao Tan, Fujun Luan, Sai Bi, Peng Wang, Jiahao Li, Zifan Shi, Kalyan Sunkavalli, Gordon Wetzstein, Zexiang Xu, and Kai Zhang.
\newblock Dmv3d: Denoising multi-view diffusion using 3d large reconstruction model.
\newblock In \emph{International Conference on Learning Representations (ICLR)}, 2024.

\bibitem[Yao et~al.(2018)Yao, Hsu, Zhu, Wu, Torralba, Freeman, and Tenenbaum]{yao20183d}
Shunyu Yao, Tzu~Ming Hsu, Jun-Yan Zhu, Jiajun Wu, Antonio Torralba, Bill Freeman, and Josh Tenenbaum.
\newblock 3d-aware scene manipulation via inverse graphics.
\newblock In \emph{Conference on Neural Information Processing Systems (NeurIPS)}, 2018.

\bibitem[Ye et~al.(2023{\natexlab{a}})Ye, Zhang, Liu, Han, and Yang]{ye2023ip}
Hu Ye, Jun Zhang, Sibo Liu, Xiao Han, and Wei Yang.
\newblock Ip-adapter: Text compatible image prompt adapter for text-to-image diffusion models.
\newblock \emph{arXiv preprint arXiv:2308.06721}, 2023{\natexlab{a}}.

\bibitem[Ye et~al.(2023{\natexlab{b}})Ye, Wang, and Wang]{ye2023featurenerf}
Jianglong Ye, Naiyan Wang, and Xiaolong Wang.
\newblock Featurenerf: Learning generalizable nerfs by distilling foundation models.
\newblock In \emph{IEEE International Conference on Computer Vision (ICCV)}, 2023{\natexlab{b}}.

\bibitem[Yenphraphai et~al.(2024)Yenphraphai, Pan, Liu, Panozzo, and Xie]{yenphraphai2024image}
Jiraphon Yenphraphai, Xichen Pan, Sainan Liu, Daniele Panozzo, and Saining Xie.
\newblock Image sculpting: Precise object editing with 3d geometry control.
\newblock In \emph{IEEE Conference on Computer Vision and Pattern Recognition (CVPR)}, 2024.

\bibitem[Yu et~al.(2021)Yu, Ye, Tancik, and Kanazawa]{yu2021pixelnerf}
Alex Yu, Vickie Ye, Matthew Tancik, and Angjoo Kanazawa.
\newblock pixelnerf: Neural radiance fields from one or few images.
\newblock In \emph{IEEE Conference on Computer Vision and Pattern Recognition (CVPR)}, 2021.

\bibitem[Yu et~al.(2022)Yu, Xu, Koh, Luong, Baid, Wang, Vasudevan, Ku, Yang, Ayan, et~al.]{yu2022scaling}
Jiahui Yu, Yuanzhong Xu, Jing~Yu Koh, Thang Luong, Gunjan Baid, Zirui Wang, Vijay Vasudevan, Alexander Ku, Yinfei Yang, Burcu~Karagol Ayan, et~al.
\newblock Scaling autoregressive models for content-rich text-to-image generation.
\newblock In \emph{International Conference on Machine Learning (ICML)}, 2022.

\bibitem[Yuan et~al.(2024)Yuan, Cao, Wang, Qi, Yuan, and Shan]{yuan2024customnet}
Ziyang Yuan, Mingdeng Cao, Xintao Wang, Zhongang Qi, Chun Yuan, and Ying Shan.
\newblock Customnet: Object customization with variable-viewpoints in text-to-image diffusion models.
\newblock In \emph{ACM Multimedia}, 2024.

\bibitem[Zhang et~al.(2022)Zhang, Ramanan, and Tulsiani]{zhang2022relpose}
Jason~Y Zhang, Deva Ramanan, and Shubham Tulsiani.
\newblock Relpose: Predicting probabilistic relative rotation for single objects in the wild.
\newblock In \emph{European Conference on Computer Vision (ECCV)}, 2022.

\bibitem[Zhang et~al.(2024)Zhang, Lin, Kumar, Yang, Ramanan, and Tulsiani]{zhang2023cameras}
Jason~Y Zhang, Amy Lin, Moneish Kumar, Tzu-Hsuan Yang, Deva Ramanan, and Shubham Tulsiani.
\newblock Cameras as rays: Sparse-view pose estimation via ray diffusion.
\newblock In \emph{International Conference on Learning Representations (ICLR)}, 2024.

\bibitem[Zhang and Agrawala(2023)]{zhang2023adding}
Lvmin Zhang and Maneesh Agrawala.
\newblock Adding conditional control to text-to-image diffusion models.
\newblock In \emph{IEEE International Conference on Computer Vision (ICCV)}, 2023.

\bibitem[Zhang et~al.(2021)Zhang, Chen, Ling, Gao, Zhang, Torralba, and Fidler]{zhang2020image}
Yuxuan Zhang, Wenzheng Chen, Huan Ling, Jun Gao, Yinan Zhang, Antonio Torralba, and Sanja Fidler.
\newblock Image gans meet differentiable rendering for inverse graphics and interpretable 3d neural rendering.
\newblock In \emph{International Conference on Learning Representations (ICLR)}, 2021.

\bibitem[Zhang et~al.(2023)Zhang, Dong, Tang, Huang, Huang, Ma, Lee, Deussen, and Xu]{zhang2023prospect}
Yuxin Zhang, Weiming Dong, Fan Tang, Nisha Huang, Haibin Huang, Chongyang Ma, Tong-Yee Lee, Oliver Deussen, and Changsheng Xu.
\newblock Prospect: Prompt spectrum for attribute-aware personalization of diffusion models.
\newblock \emph{ACM Transactions on Graphics (TOG)}, 2023.

\bibitem[Zhou et~al.(2022)Zhou, Girdhar, Joulin, Kr{\"a}henb{\"u}hl, and Misra]{zhou2022detecting}
Xingyi Zhou, Rohit Girdhar, Armand Joulin, Philipp Kr{\"a}henb{\"u}hl, and Ishan Misra.
\newblock Detecting twenty-thousand classes using image-level supervision.
\newblock In \emph{European Conference on Computer Vision (ECCV)}, 2022.

\bibitem[Zhou and Tulsiani(2023)]{zhou2023sparsefusion}
Zhizhuo Zhou and Shubham Tulsiani.
\newblock Sparsefusion: Distilling view-conditioned diffusion for 3d reconstruction.
\newblock In \emph{IEEE Conference on Computer Vision and Pattern Recognition (CVPR)}, 2023.

\end{thebibliography}
}

\clearpage
\appendix
\renewcommand{\thefootnote}{\arabic{footnote}}

\clearpage
\noindent{\Large\bf Appendix}
\vspace{5pt}

In \refapp{results_ablate} and \refsec{results2}, we show more ablation and results. In \refapp{eval_details}, we provide details regarding the evaluation and human preference study. Finally, in \refapp{details}, we describe all the implementation details of our method and baselines.  

\section{Ablation}\lblsec{results_ablate}

\myparagraph{Using predicted masks for background losses.} In our main paper, we calculated the background losses (\refeq{loss_background}) using ground truth masks from the dataset during training. We show in \reftbl{ablation_2} that replacing these with predicted masks results in similar performance. We use Detic~\cite{zhou2022detecting} with Segment Anything~\cite{kirillov2023segment} to predict the object mask given the object category, such as car and teddy bear.

\myparagraph{Varying number of views.} In all our experiments, we use $\sim 50$ multi-view images and their ground truth poses for training. Here, we vary the number of multi-view images to $35$ and $20$, respectively, and show its results in \reftbl{ablation_2}. As the number of views decreases, text alignment remains similar, but the accuracy of camera pose and image alignment gradually decreases. Example generations and their comparison to our main method are shown in \reffig{vary_view}.

\begin{figure}[!t]    \includegraphics[width=\linewidth]{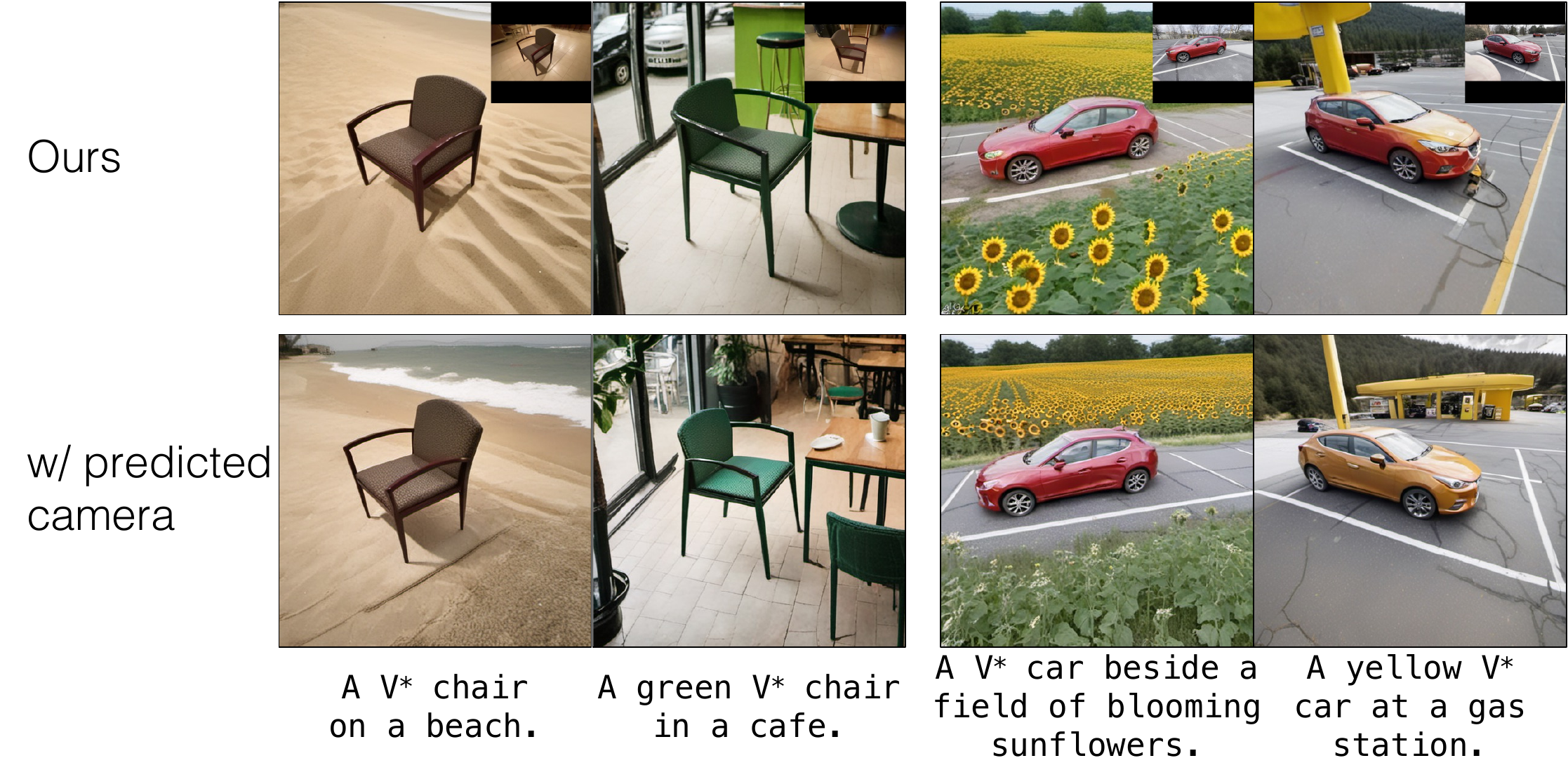} 
    \vspace{-20pt}
    \caption{\textbf{Training with predicted camera viewpoints.} We observe similar performance with predicted cameras using $\smallsim 50$ training images. }
\label{fig:pred_camera}
\end{figure}

\myparagraph{Using predicted camera viewpoints.} \nupur{Here, we use COLMAP to predict the camera viewpoints using only the $\smallsim50$ training images. Though it requires longer training, $2000$ steps compared to $1600$ for our method, the final performance is comparable to our original method. The CLIP and DINO scores remain similar while the mean angular error increases slightly to $17.12$ compared to $16.14$ of our final method on $10$ objects (COLMAP fails to run on $2$ car objects). Qualitative samples and their comparison to our final method are shown in \reffig{pred_camera}. }

\section{More Results}\lblsec{results2}

\begin{figure*}[!t]
    \begin{tabular}{cc}
    \includegraphics[width=\linewidth]{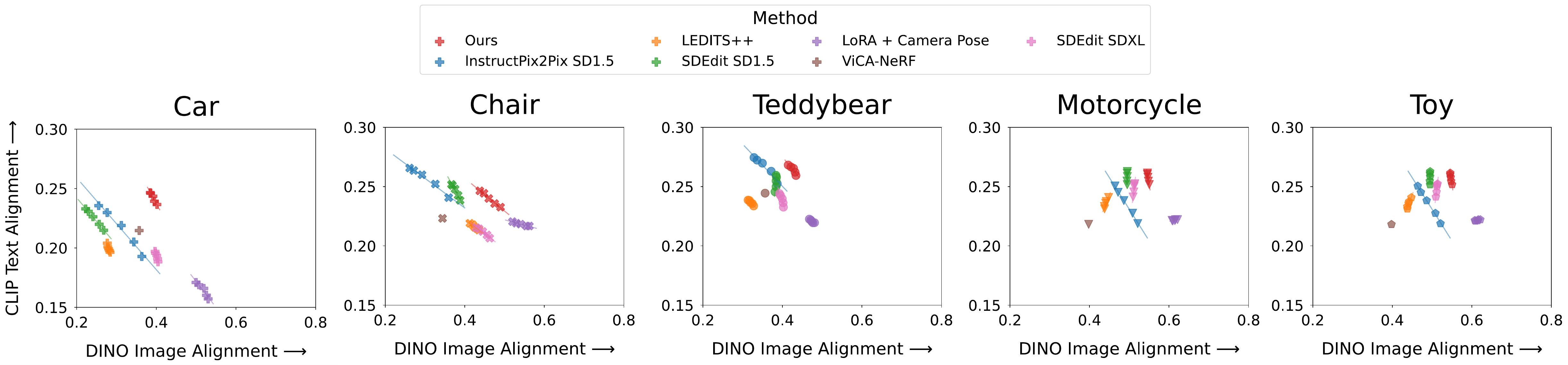}  
    \end{tabular}
    \caption{\textbf{CLIP vs. DINO scores, separated by object category}. For each category, our method achieves higher or the same text alignment compared to baselines while having on-par image alignment.)}
\label{fig:dino_clip_per_category}
\end{figure*}

\myparagraph{Per category comparison.} We show text-alignment and image-alignment scores of our method and baselines for each category in \reffig{dino_clip_per_category} while varying the text guidance scale for each method from $5.0$ to $10.0$, except for LEDITS++~\cite{brack2023ledits++} which recommends varying the concept editing guidance scale from $10.0$ to $15.0$. For each guidance scale, we generate $288$ images. Our method usually results in a higher CLIP score than the baselines, indicating better text alignment. For all categories, we show the linear fit curve over the different guidance scales, and our method lies at the Pareto-frontal compared to the baseline methods, either performing similarly or better.

\myparagraph{Comparison with SDEdit+SDXL.} We add a comparison to SDEdit with the Stable Diffusion-XL model here, which we found to perform worse or on par with the Stable Diffusion-1.5 version of the model.
\reffig{dino_clip_per_category} shows the quantitative comparison.

\begin{figure*}[!t]
    \centering
    \includegraphics[width=0.9\linewidth]{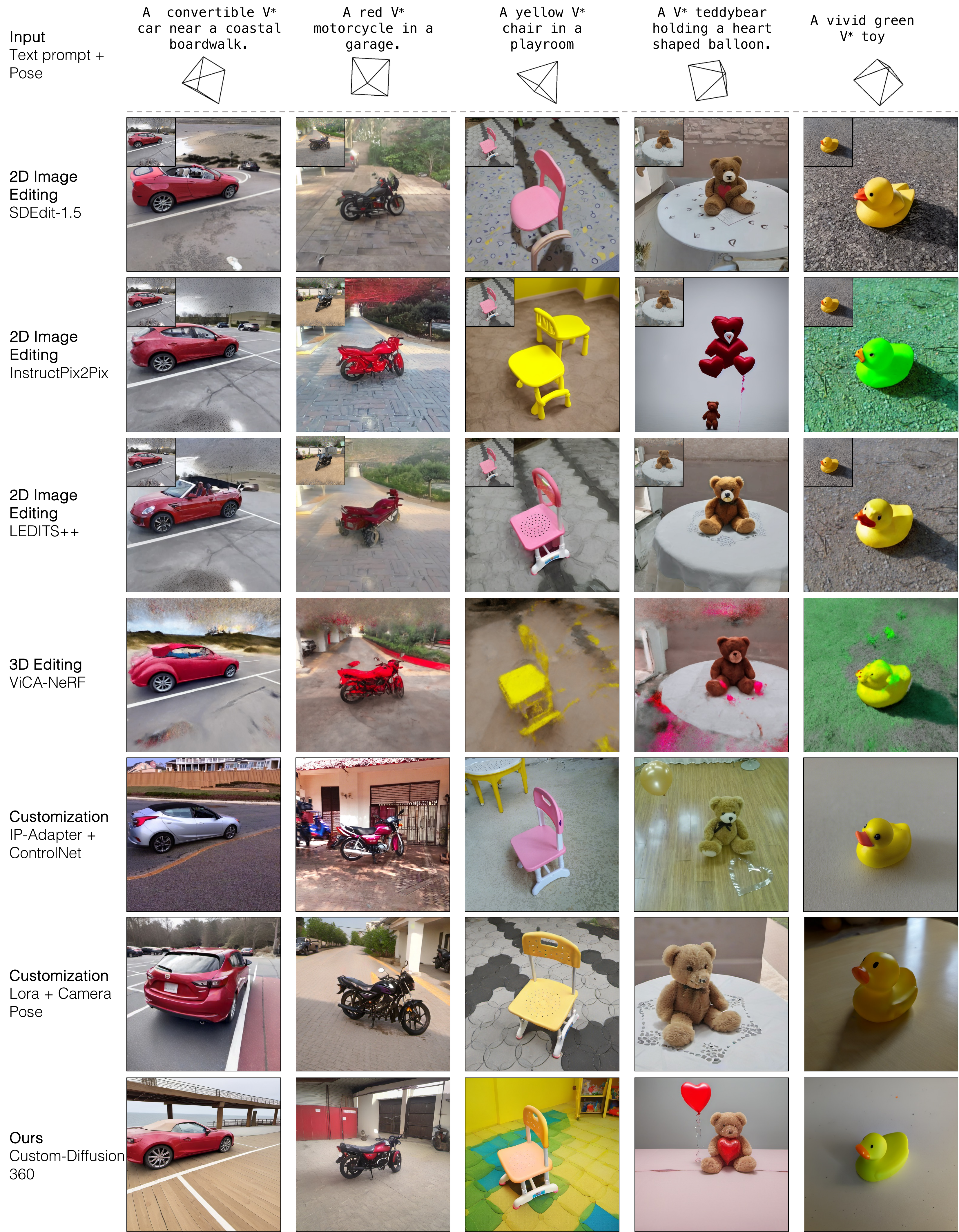}
    \caption{Additional qualitative comparison of our method with baselines, given various prompts and target pose conditions.}
\label{fig:result_appendix}
\end{figure*}

\myparagraph{Comparison to IP-Adapter + ControlNet.} \nupur{Here, we compare against IP-Adapter~\cite{ye2023ip} + depth ControlNet~\cite{zhang2023adding} for object viewpoint control. Given a target viewpoint, we render the image using a trained NeRF model of the object and use its predicted depth as a condition in depth ControlNet. We select the recommended hyperparameters that also worked best qualitatively, i.e., $0.6$ scale for the IP-Adapter and $0.5$ as ControlNet conditioning scale. The CLIP and DINO scores for the baseline are $0.251$ and $0.497$, respectively, similar to our method's CLIP and DINO  scores of $0.253$ and $0.481$. But in pairwise user-study comparison, our method is preferred by $60 \pm 3\%$ and $75.67 \pm 2.6\%$ for text-alignment and image-alignment. Qualitatively, we also observe that applying multiple conditionings, including depth and image, leads to lower text alignment as shown in \reffig{result_appendix} ($5^{\text{th}}$ row). }

\myparagraph{Comparison to LoRA.} We compare our method to the customization method LoRA~\cite{hu2021lora, loraimplementation} on the text and image alignment metrics here. The CLIP and DINO scores for LoRA are $0.275$ and $0.468$, respectively. In comparison, our CLIP and DINO scores are $0.253$ and $0.481$, respectively. Though the CLIP score is marginally lower, our method performs better in preserving the target concept while allowing additional object viewpoint control for the custom object. 

\begin{figure*}[!t]
    \centering
    \includegraphics[width=\linewidth]{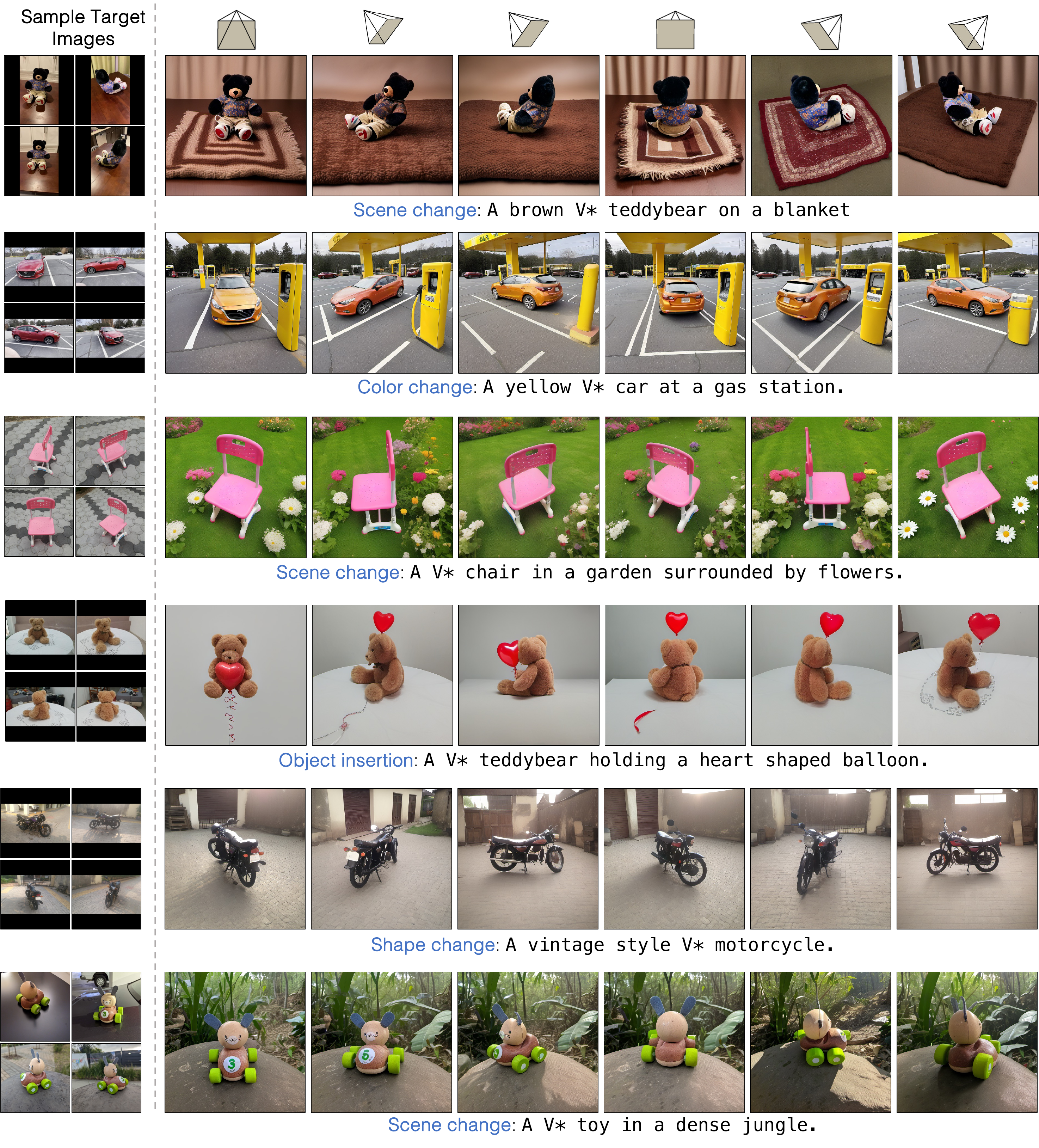}
    \caption{Additional qualitative samples of our method while varying the camera pose condition for the custom object.}
\label{fig:result2_appendix}
\end{figure*}

\myparagraph{Qualitative comparison.} We show more qualitative comparisons between our method and baselines in \reffig{result_appendix}. \reffig{result2_appendix} shows more samples from varying camera poses for our method.

\begin{table}[!t]
\centering
\setlength{\tabcolsep}{5pt}
\resizebox{\linewidth}{!}{
\begin{tabular}{l c c  c c c}
\toprule
\textbf{Method}
& \shortstack[c]{\textbf{Text} \textbf{Alignment}}
& \multicolumn{1}{@{} c}{\shortstack[c]{\textbf{Image} \textbf{Alignment}}} & \multicolumn{2}{@{} c}{\shortstack[c]{\textbf{Camera-pose} \textbf{Accuracy}}}\\
 & CLIPScore$\uparrow$ & DINO-v2$\uparrow$  & {\shortstack[c]{ Angular\\ error $\downarrow$}} & {\shortstack[c]{Camera center \\ error $\downarrow$}}  \\
\midrule
\textbf{Ours}  &   0.248	&  0.471  & 14.19 &  0.080 \\
\textbf{w/ predicted mask}   & 0.246 & \textbf{0.475} & 13.80 &  0.086  \\
\textbf{w/ 35 views}   & 0.250 & 0.439 &  18.09	&  0.108  \\
\textbf{w/ 20 views}   & \textbf{0.254} & 0.448 &  18.96	&  0.108  \\
\bottomrule
\vspace{-2pt}
\end{tabular}
}
\vspace{-8pt}
\caption{\textbf{Results with predicted masks and variable views.} Our method works similarly even when using predicted masks instead of ground truth masks for mask-based losses. When varying the number of views, the performance gradually drops w.r.t. camera-pose accuracy and image alignment. 
}

\label{tbl:ablation_2}
\vspace{-10pt}
\end{table}

\section{Evaluation.}\lblsec{eval_details}

\myparagraph{Evaluation text prompts.} As mentioned in the main paper, we used ChatGPT to generate text prompts for evaluating our method and baselines. \reftbl{prompts_eval} lists all of the prompts. An example instruction given to ChatGPT to generate prompts for the motorcycle category is: ``Provide 16 diverse captions for plausible naturally occurring images of a motorcycle.
Only follow one of the options given below while generating the captions. 
Thus, four captions per option. Each caption should have a simple sentence structure.
(1) change the background scene of the motorcycle,
(2) insert a new object in the scene with the motorcycle
(3) change the type of the  motorcycle
(4) change the color of the motorcycle''. 

\begin{figure*}[!t]
    \includegraphics[width=\linewidth]{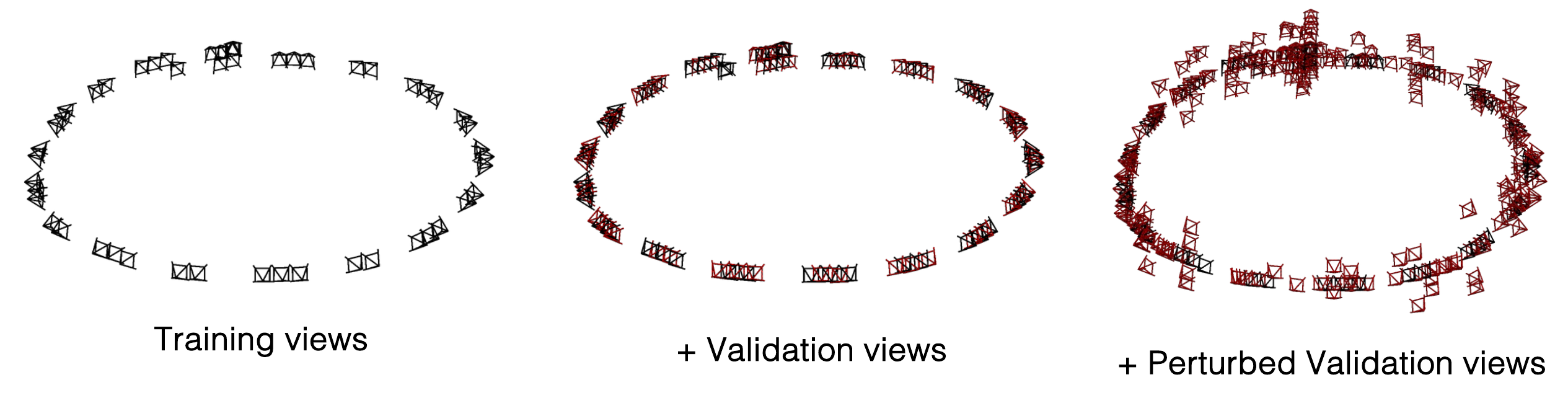} 
    \vspace{-20pt}
    \caption{\textbf{Sample training and validation (+perturbed) views.} We show sample training and validation views that we use in our method. For quantitative evaluation, we perturb the location and focal length of the validation camera poses to create the final set of evaluation target camera poses. }
\label{fig:views_vis}
\end{figure*}

\myparagraph{Evaluation camera pose.} \reffig{views_vis} shows sample training and validation camera poses for the car object. To measure the camera pose accuracy, we randomly select $6$ validation camera poses (\reffig{views_vis}, $2^{\text{nd}}$ column) and generate images using one of the $16$ text prompts. We then use RayDiffusion~\cite{zhang2023cameras} to predict the camera pose from the $6$ generated images and calculate its error with the target camera pose. The validation camera poses are such that the principal axis of the camera points towards the object. For text- and image-alignment metrics, we use the perturbed validation camera poses (\reffig{views_vis}, $3^{\text{rd}}$ column).

\myparagraph{Human preference study.} We perform a pairwise comparison of our method with each baseline. To measure text alignment, we show the input text prompt and the two images generated by ours  and the baseline method, respectively, and ask: ``Which image is more consistent with the following text?'' For the image alignment, we show $3$-$4$ target images at the top and ask: ``Which of the below images is more consistent with the shown target object?'' To measure photorealism, we only show the two generated images with the question: ``Which of the below images is more photorealistic?''

\myparagraph{DINO image alignment metrics.} We use DINOv2~\cite{oquab2023dinov2} as the pretrained model to measure image alignment. For each generated image, we measure its mean similarity to all the training images of the target concept. We crop the object region in the training images using masks to measure the similarity only with the target concept. 

\section{Implementation Details}\lblsec{details}

\subsection{Our Method}

We fine-tune a pretrained Stable Diffusion-XL $32$-bit floating point model with a batch size of $4$. We update the new parameters with a learning rate of $1 \times 10^{-4}$. During training, we bias the sampling of time towards later time steps~\cite{mou2024t2i} since pose information is more crucial in the early stages of denoising. Training is done for $1600$ iterations, which takes $\sim 45$ minutes on $4$ A100-GPUs with $40GB$ VRAM. 
At each training step, we sample $5$ views (maximum possible in GPU memory) equidistant from each other and use the first as the target viewpoint and the others as references. We crop the reference images tightly around the object bounding box and modify the camera intrinsics accordingly. We modify $12$ transformer layers with pose-conditioning out of a total of $70$ transformer layers in Stable Diffusion-XL, with $4$ in the encoder, $3$ in the middle, and $5$ in the decoder blocks of the U-Net. Further, in a particular encoder or decoder block, we use the density predicted by previous FeatureNeRF blocks to importance sample the points along the ray for the next FeatureNeRF block $90\%$ of the times. This improves the performance on concepts with thin structuress like chairs. For rendering, we sample $24$ points along the ray. The training hyperparameters, $\lambda_{\text{rgb}}$, $\lambda_{\text{s}}$, $\lambda_{\text{bg}}$ are set to the to $5$, $10$, and $10$ for all experiments.

During inference, we use an image guidance scale of $3.5$ and a text guidance scale of $7.5$ in \refeq{inference_ours}. All images for evaluation are generated with $50$ sampling steps using the default Euler scheduler~\cite{karras2022elucidating}. 
With these settings, the wallclock runtime to generate one image given $8$ reference views is $10$ seconds. We cache the reference view features for each customized model. For comparison, it takes $6$ seconds to generate an image given no reference views.

\begin{figure}[!t]    \includegraphics[width=\linewidth]{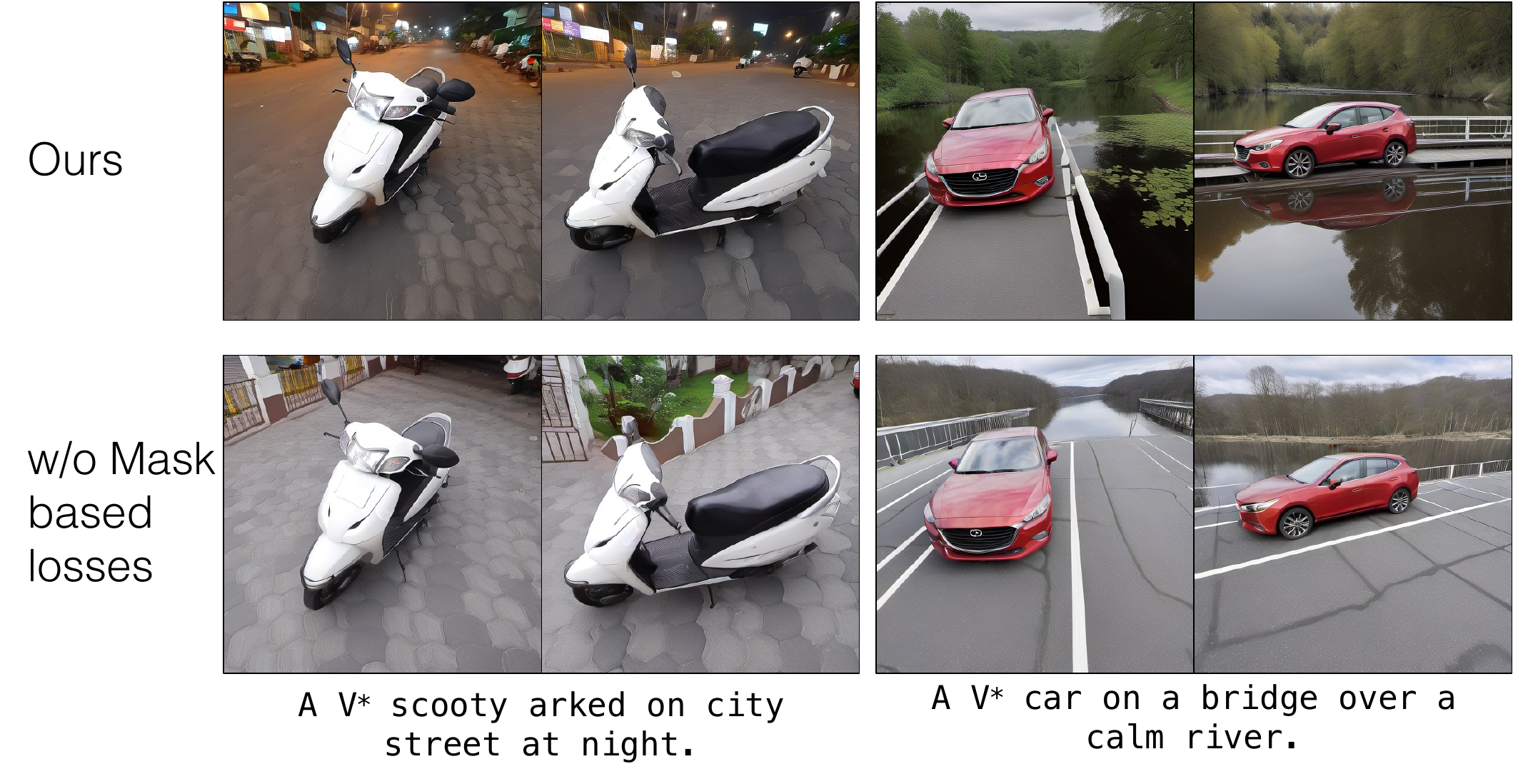} 
    \vspace{-20pt}
    \caption{\textbf{Role of mask-based loss in FeatureNeRF.}  Not having silhouette and background losses results in the model overfitting to the background features of the training image, e.g., the trees in the background.}
\label{fig:mask_loss}
\end{figure}

\begin{figure}[!t]    \includegraphics[width=\linewidth]{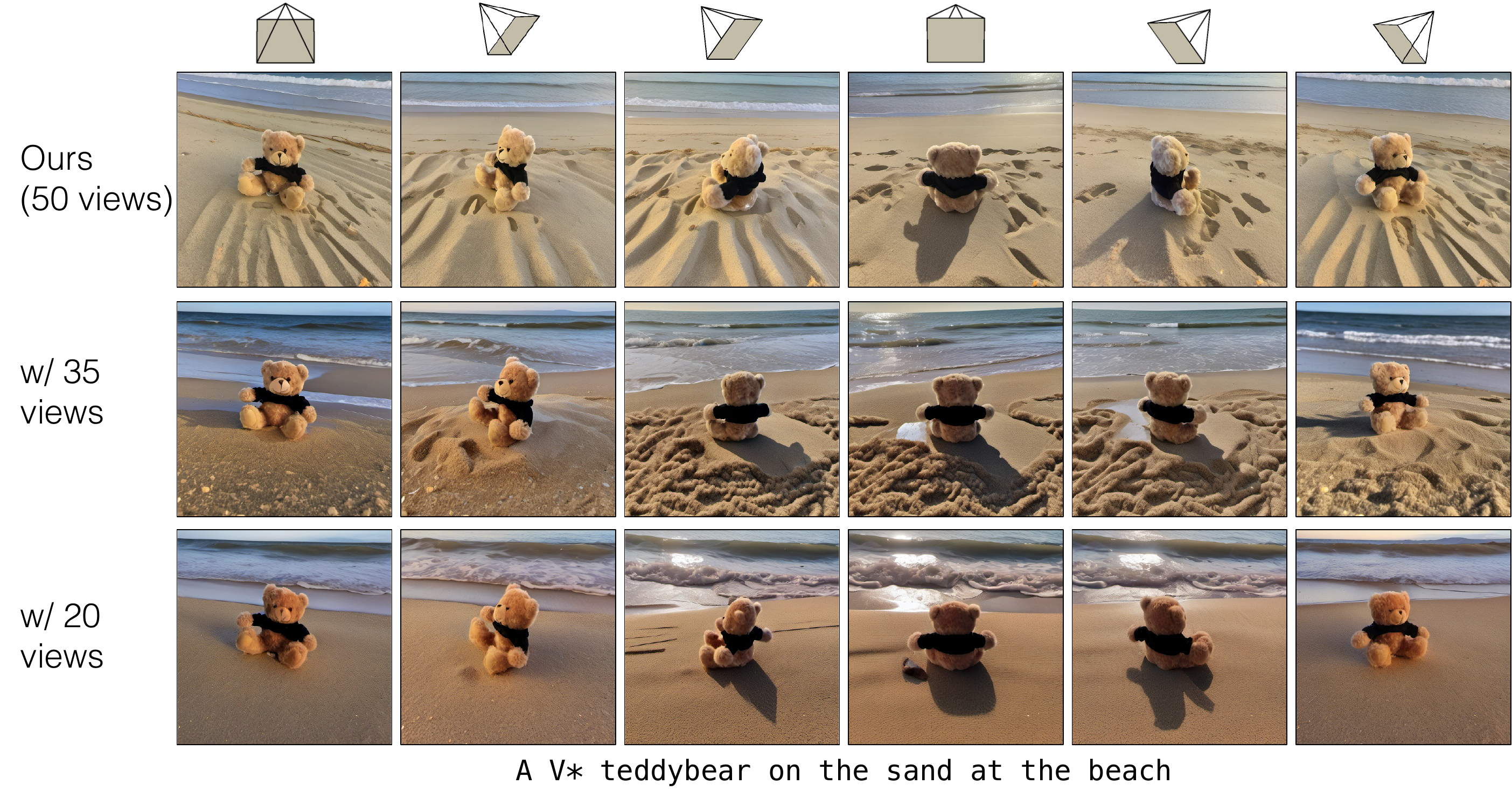} 
    \vspace{-20pt}
    \caption{\textbf{Number of training views.} As we decrease the number of training views, some target camera poses are not respected in the output generations, specifically back-facing viewpoints. We observe that the model has a bias towards generating front-facing objects. }
\label{fig:vary_view}
\end{figure}

\subsection{Baselines}
For 2D image editing methods, we first render the image given the target camera pose using a trained NeRF model~\cite{tancik2023nerfstudio}. To remove floater artifacts outside of the object mask in the rendered image, we run a pre-processing step using the SDEdit Stable Diffusion-XL denoising ensemble \cite{podell2023sdxl} with negative prompts ``blurry, blur''. The inpainted image is then passed to the 2D image editing method as the input. For each baseline, we calculate metrics at $5$ different guidance scales and keep the remaining hyperparameters the same across all object categories as much as possible.

\myparagraph{SDEdit~\cite{meng2021sdedit}.} This uses the forward process to create a noisy image at some intermediate timestep and runs the backward denoising process with the new text prompt. For the optimal strength value, we ran a grid search from $0.5$ to $0.8$. We report our evaluation metrics on the best-performing strength of $0.5$ (highest average text alignment to image alignment ratio). We run all inference in float$16$. For SDEdit with Stable Diffusion-1.5, we use the recommended PNDM sampler~\cite{liu2022pseudo} and set the number of inference steps to $50$. With a strength of $0.5$, the number of denoising steps run is $25$ and takes $1$ second. In the case of Stable Diffusion-XL, we apply the base model and refiner model as an ensemble of expert denoisers~\cite{podell2023sdxl}. We run the base model with the default Euler scheduler~\cite{karras2022elucidating} for $15$ steps at strength $0.5$ and the refiner for $5$ more steps. With these settings, the wallclock runtime to generate one image is $5$ seconds.

\myparagraph{LEDITS++~\cite{brack2023ledits++}.} This is a more recent method that proposes a new inversion technique to embed the image in latent space. It then constructs semantically grounded masks from the U-Net's cross-attention layers and noise estimates to constrain the edit regions corresponding to the new text prompt. The user must adjust the LEDITS++ hyperparameter values according to whether they wish to modify large or small regions of the image, e.g., background changes vs. object appearance edits. To find the best hyperparameters, we refer to the recommended values provided in the official implementation. Thus, for prompts that edit the object's appearance, we keep the target prompt guidance scale at $8$, the edited concept's threshold at $0.9$, the default number of inference steps at $50$, and the proportion of initial denoising steps skipped to $0.25$. For prompts that change the background significantly, we additionally change the target prompt from the original image's BLIP caption~\cite{li2023blip} to the background-changing prompt. For example, the edit in \reffig{vary_view} is achieved in LEDITS++ by changing the target prompt from ``A teddy bear sitting on a table in a living room'' to ``A teddybear on the sand at the beach''. On the other hand, to edit the color of the teddy bear to gray using LEDITS++, the target prompt would still be ``A teddy bear sitting on a table in a living room'' but the concept ``A gray teddybear'' would be added to guided the generated image in that direction.

\myparagraph{InstructPix2Pix~\cite{brooks2023instructpix2pix}.} This image-editing technique trains a model to follow editing instructions and can edit a new input image in a feedforward manner. We use the official model released with the paper based on Stable Diffusion-1.5, set the image guidance scale to the suggested value of $1.5$,
and use the default Euler scheduler with $50$ inference steps.

\myparagraph{ViCA-NeRF~\cite{dong2024vica}.} This is a 3D editing method that provides improved multiview consistency and training speed compared to Instruct-NeRF2NeRF~\cite{haque2023instruct}. First, the editing stage edits key views from the set of multiview images using InstructPix2Pix, reprojects the editing results to other views using the camera pose and depth information, and blends the edits in the latent diffusion model's feature space. Second, the NeRF training stage uses the edited multiview images to optimize the unedited NeRF and produce a 3D-edited NeRF. We use the official implementation's hyperparameters of text guidance scale $7.5$ and image guidance scale $1.5$. For concepts from the NAVI dataset, we select images from a single scene. For our method, we use images from multiple scenes since we only model the foreground object.

\myparagraph{LoRA~\cite{hu2021lora}.} This is a 2D customization method that trains low-rank adapters in linear layers of the diffusion model U-Net when fine-tuning on the new custom concept. We fine-tune Stable Diffusion-XL using the LoRA fine-tuning technique with rank $64$ adapters added to all the linear layers in the attention blocks. We use the recommended learning rate of $1\times 10^{-4}$ with batch-size $16$ and train for $1000$ iterations. We sample the regularization images as in our method $25\%$ of the times. For training, we use the text prompt {\menlo photo of a V$^*$ \{category\} }, where {\menlo V$^*$} is a fixed token.

\myparagraph{LoRA + Camera pose.} We modify the LoRA customization method to include the camera pose condition and text prompt. To achieve this, in every cross-attention layer, we concatenate the flattened camera projection matrix with the text transformer output along the feature dimension and use a one-layer MLP to project it back to the original dimension. The one-layer MLP is trained along with LoRA adapter modules, with all other hyperparameters kept the same as the above LoRA baseline. To match our method, we also biased the sampling of time towards later time steps. Like ours, the reconstruction loss is only applied in the masked region. We train the model for a total of $2000$ iterations.

\section{Change log}
\myparagraph{v1:} Original draft.

\myparagraph{v2:} Aditional comparison to 3D-editing based methods in \refsec{results1} and more ablation experiments in \refapp{results_ablate}. Updated citations and writing.

\begin{table}[!t]
    \setlength{\tabcolsep}{5pt}
    \resizebox{0.85\linewidth}{!}{
    \begin{tabular}{l l}
    \toprule
    \shortstack[c]{\textbf{Object Category} } 
    & \shortstack[c]{\textbf{Evaluation Prompt} } \\
    \midrule
     \multirow{16}{*}{\shortstack[c]{Car }} &  A car parked by a snowy mountain range. \\
     & A car on a bridge over a calm river. \\
     & A car in front of an old, brick train station. \\
     & A car beside a field of blooming sunflowers. \\
     & A car with a bike rack on top. \\
     & A car next to a picnic table in a park. \\
     & A car with a guitar leaning against it. \\
     & A car with a kayak mounted on the roof.\\
     & A minivan car outside a school, during pickup time.\\
     & A convertible car near a coastal boardwalk. \\
     & A jeep car on a rugged dirt road. \\
     & A volkswagon beetle car in front of a luxury hotel. \\
     & A red car in a mall parking lot. \\
     & A yellow car at a gas station. \\
     & A green car in a driveway, next to a house. \\
     & A black car in a busy city street. \\
      \midrule
     \multirow{16}{*}{\shortstack[c]{Chair}} &  A chair on a balcony overlooking the city skyline. \\
    & A chair in a garden surrounded by flowers. \\
    & A chair on a beach. \\
    & A chair in a library next to a bookshelf. \\
    & A chair with a plush toy sitting on it. \\
    & A chair beside a guitar on a stand. \\
    & A chair next to a potted plant. \\
    & A chair with a colorful cushion on it. \\
    & A rocking chair on a porch. \\
    & An office chair in a home study. \\
    & A folding chair at a camping site. \\
    & A high chair in a kitchen. \\
    & A red chair in a white room. \\
    & A black chair in a classroom. \\
    & A green chair in a café. \\
    & A yellow chair in a playroom. \\
    \midrule
    \multirow{16}{*}{\shortstack[c]{Teddybear}}  & A teddybear on a park bench under trees. \\
    & A teddybear at a window with raindrops outside. \\
    & A teddybear on the sand at the beach. \\
    & A teddybear on a cozy armchair by a fireplace. \\
    & A teddybear with a stack of children's books on the side. \\
    & A teddybear next to a birthday cake with candles. \\
    & A teddybear with a small toy car. \\
    & A teddybear holding a heart-shaped balloon. \\
    & A teddybear in a pink barbie costume. \\
    & A large teddybear in a batman costume. \\
    & A teddybear dressed as a construction worker. \\
    & A teddybear in a superhero costume. \\
    & A pink teddybear on a shelf. \\
   &  A brown teddybear on a blanket. \\
    & A white teddybear. \\
    & A gray teddybear. \\
    \midrule
     \multirow{16}{*}{\shortstack[c]{Motorcycle}}  &  A motorcycle parked on a city street at night. \\
    &  A motorcycle beside a calm lake. \\
    &  A motorcycle on a mountain road with a scenic view. \\
    &  A motorcycle in front of a graffiti-covered urban wall. \\
    &  A motorcycle with a guitar strapped to the back. \\
    &  A motorcycle next to a camping tent. \\
    &  A golden retriever riding motorcycle. \\
    &  A cat riding motorcycle. \\
    &  A cruiser motorcycle in a parking lot. \\
    &  A scooter like motorcycle. \\
    &  A vintage style motorcycle. \\
    &  A dirt bike motorcycle on a trail in the woods. \\
    &  A red motorcycle in a garage. \\
    &  A green motorcycle. \\
    &  A blue motorcycle. \\
    &  A silver motorcycle.\\
    \midrule
    \multirow{16}{*}{\shortstack[c]{Toy }} & Toy on a sandy beach, with waves crashing in the background \\ 
& A toy sitting in a grassy field, surrounded by wildflowers. \\
& A toy on a rocky mountain top, overlooking the valley below.\\
& A toy in a dense jungle.\\
& A toy with a tiny book placed beside it on a wooden table.\\
& A toy floating next to a colorful beach ball in a bathtub.\\
& A toy with a small globe resting next to it.\\
& A toy and an umbrella in a cozy living room.\\
& A plush toy on a sunny windowsill.\\
& A wooden toy.\\
& An origami of toy.\\
& A clay figurine  of toy.\\
& A bright red toy.\\
& A deep blue toy on a bed.\\
& A vivid green toy. \\
& A neon pink toy. \\
     \bottomrule
    \vspace{-10pt}
    \end{tabular}
    }
    \vspace{-2pt}
    \caption{\textbf{Evaluation prompts.} Here, we list the final prompts that were used for evaluation for the car, chair, teddy bear, motorcycle, and toy categories. }
    
    \label{tbl:prompts_eval}
    \vspace{-6pt}
\end{table}

\begin{figure}[!t]
    \centering
    \includegraphics[width=\linewidth]{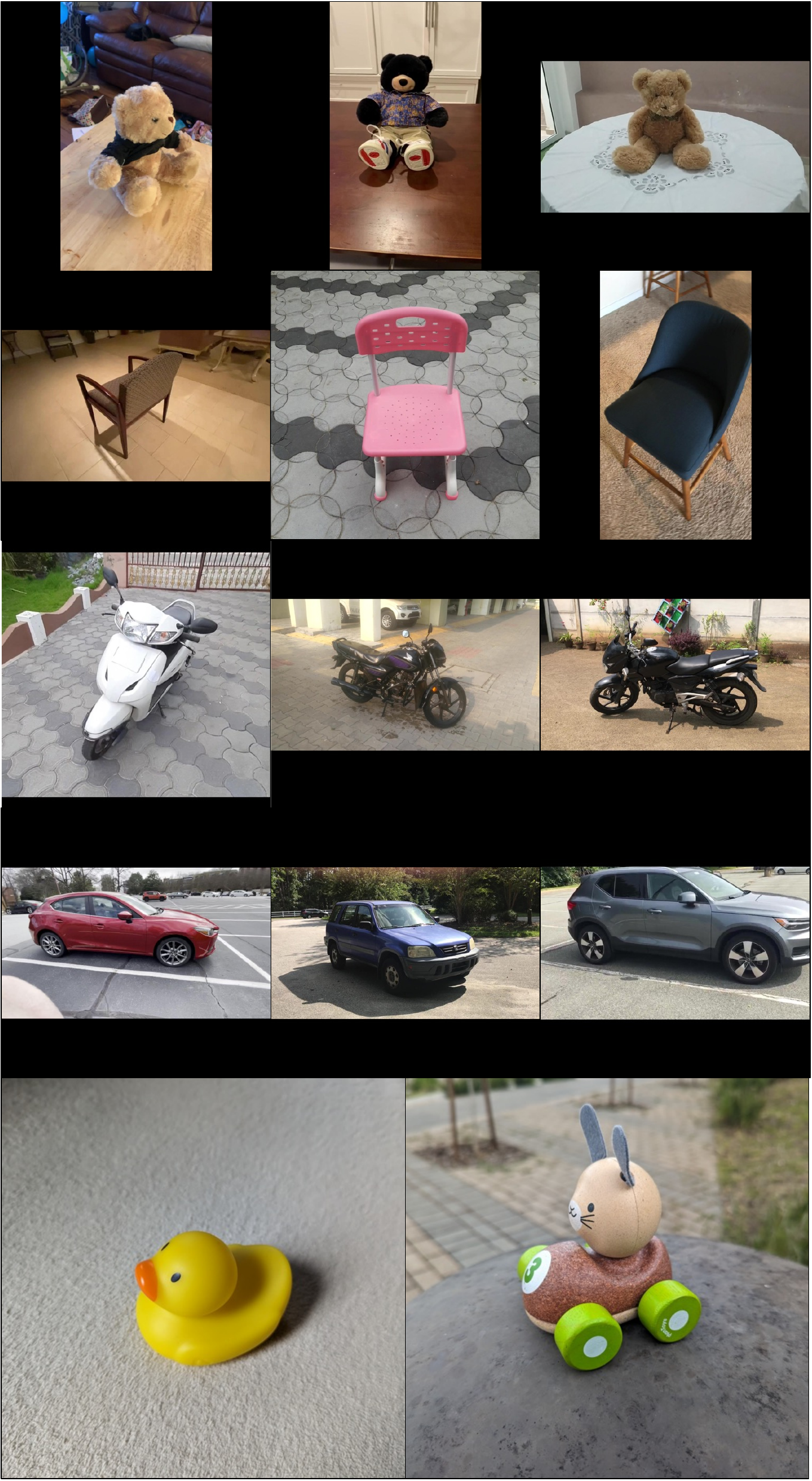}
    \caption{\textbf{Target concepts}. Sample images from the $14$ target concepts used as the dataset for evaluating our method.}
\label{fig:sampledata}
\end{figure}

\clearpage

\end{document}